\documentclass[journal]{IEEEtran}

\usepackage[export]{adjustbox}
\usepackage{booktabs}
\usepackage{graphicx}
\usepackage{tikz}
\usepackage[outline]{contour}
\contourlength{1.5pt}
\graphicspath{{./figs/}}

\usepackage{cite}
\usepackage{todonotes}

\usepackage{tabularx}
\usepackage{siunitx}

\usepackage{algorithm}
\usepackage{algpseudocode}
\makeatletter
\let\OldStatex\Statex
\renewcommand{\Statex}[1][0]{%
  \setlength\@tempdima{\algorithmicindent}%
  \OldStatex\hskip\dimexpr#1\@tempdima\relax}
\makeatother

\usepackage{etoolbox}
\makeatletter
\patchcmd{\@makecaption}
  {\scshape}
  {}
  {}
  {}
\makeatletter
\patchcmd{\@makecaption}
  {\\}
  {.\ }
  {}
  {}
\makeatother
\def\tablename{Table}

\def\Mw{{\mathbf M}}

\def\Dw{{\mathbf D}}

\usepackage[cmex10]{amsmath}
\usepackage{amssymb, amsthm}

\newtheorem{problem}{Problem}

\DeclareMathOperator*{\argmin}{argmin}

\usepackage{stfloats}
\usepackage{url}

\begin{document}
\title{Void Filling of Digital Elevation Models\\ with Deep Generative Models}
%
%

\author{Konstantinos Gavriil,
        Georg Muntingh
        and~Oliver~J. D. Barrowclough
\thanks{K. Gavriil is with Evolute GmbH and the Institute of Discrete Mathematics and Geometry, Vienna University of Technology, Wiedner Hauptstrasse 8-10/104, A-1040 Vienna, Austria (email: gavriil@evolute.at).}%

\thanks{G. Muntingh and O.J.D. Barrowclough are with SINTEF Digital, Forskningsveien 1, 0373 Oslo, Norway (email: georg.muntingh@sintef.no and oliver.barrowclough@sintef.no)}%

\thanks{This project has received funding from the European Union's Horizon 2020 research and innovation programme under the Marie Sk\l{}odowska-Curie grant agreement No 675789. This projected was also supported by an IKTPLUSS grant, project number 270922, from the Research Council of Norway.}%

\thanks{Manuscript received November 30, 2018; revised February 4, 2019; accepted February 22, 2019.}}

\markboth{IEEE GEOSCIENCE AND REMOTE SENSING LETTERS}%
{Shell \MakeLowercase{\textit{et al.}}: Bare Demo of IEEEtran.cls for Journals}


\maketitle

\begin{abstract}
In recent years, advances in machine learning algorithms, cheap computational resources, and the availability of big data have spurred the deep learning revolution in various application domains. In particular, supervised learning techniques in image analysis have led to superhuman performance in various tasks, such as classification, localization, and segmentation, while unsupervised learning techniques based on increasingly advanced generative models have been applied to generate high-resolution synthetic images indistinguishable from real images.

In this paper we consider a state-of-the-art machine learning model for image inpainting, namely a Wasserstein Generative Adversarial Network based on a fully convolutional architecture with a contextual attention mechanism. We show that this model can successfully be transferred to the setting of digital elevation models (DEMs) for the purpose of generating semantically plausible data for filling voids. Training, testing and experimentation is done on GeoTIFF data from various regions in Norway, made openly available by the Norwegian Mapping Authority.

\end{abstract}

\begin{IEEEkeywords}
Digital elevation models; unsupervised learning; predictive models; remote sensing
\end{IEEEkeywords}

%
\IEEEpeerreviewmaketitle

\section{Introduction}

\IEEEPARstart{I}{n the} field of remote sensing and data capture, a common issue is that certain areas are not completely or adequately covered, resulting in regions of `missing data'.
The reasons behind this issue vary depending in the data capture technique applied.
For example, NASA's Shuttle Radar Topography Mission (SRTM) from the early 2000s attempted to provide a complete digital elevation model (DEM) of most of the globe.
However, issues with missing data arose in regions of high gradient, such as mountainous regions, meaning very rugged terrain was often not well captured \cite{Luedeling2007}.
In stereophotogrammetry, where pairs of aerial or satellite images are matched to create digital elevation models, failures can occur when there are differences between the content of two images (e.g. variable cloud cover at different capture times) or in regions where there are not enough features to perform a successful matching.
In light detection and ranging (LIDAR) capture, the sensors are typically positioned together with the source of illumination.
This means that data is only captured on the `visible' surface and no data is captured on the `back side' of objects unless the sensor is moved.


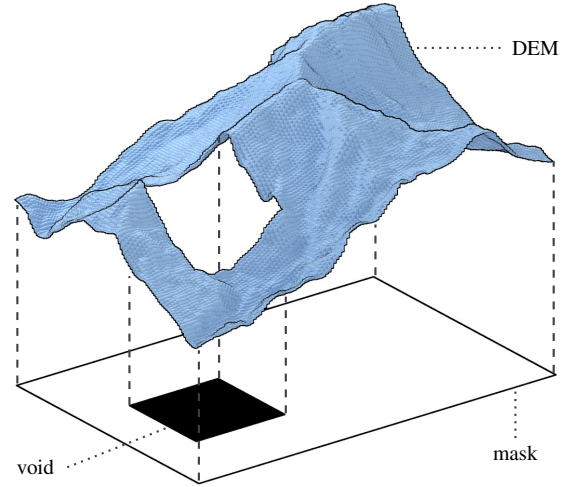
\begin{figure}[!t]
\centering
\footnotesize
\begin{tikzpicture}
  \node[anchor=south west,inner sep=0] (image) at (0,0) {\includegraphics[width=0.4\textwidth]{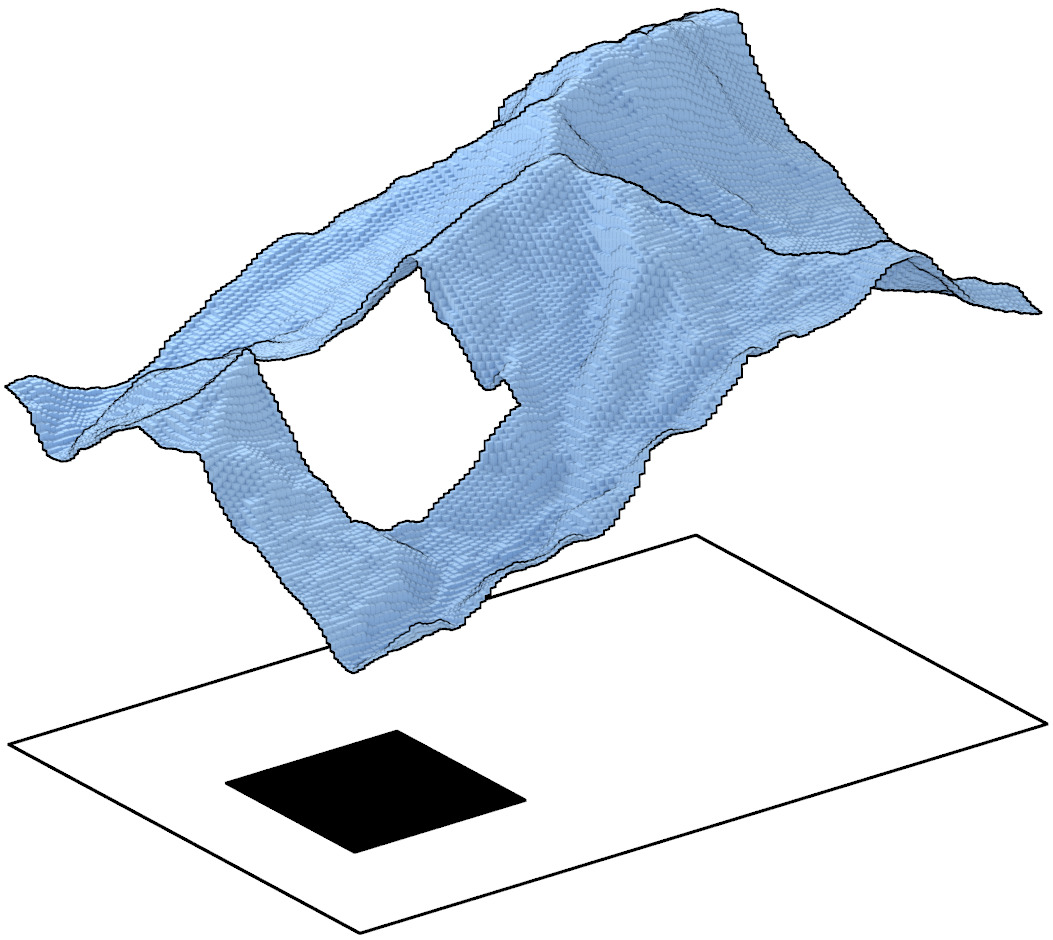}};
  \begin{scope}[x={(image.south east)},y={(image.north west)}]

  \node[right] at (0.9,0.9) {DEM};
  \draw[dotted, thick, darkgray](0.74,0.9) -- (0.9,0.9);
  \node[below] at (0.92,0.1) {mask};
  \draw[dotted, thick, darkgray](0.92,0.1) -- (0.92,0.198);
  \node[left] at (0.09,0.04) {void};
  \draw[dotted, thick, darkgray](0.28,0.12) -- (0.1,0.04);

  \draw[dashed,thick,-, darkgray](0.009,0.212) -- (0.009,0.578);
  \draw[dashed,thick,-, darkgray](0.341,0.013) -- (0.341,0.28);
  \draw[dashed,thick,-, darkgray](0.99,0.234) -- (0.99,0.668);
  \draw[dashed,thick,-, darkgray](0.664,0.434) -- (0.664,0.538);

  \draw[dashed,thick,-, darkgray](0.214,0.168) -- (0.214,0.432);
  \draw[dashed,thick,-, darkgray](0.5,0.151) -- (0.5,0.39);
  \draw[dashed,thick,-, darkgray](0.378,0.222) -- (0.378,0.303);
  \draw[dashed,thick,-, darkgray](0.378,0.445) -- (0.378,0.72);

  \end{scope}
\end{tikzpicture}

\vspace{5pt}
\caption{Visual representation of a digital elevation model, and the respective binary mask of the known and unknown values.\label{fig:dem}}
\end{figure}


Traditional methods to rectify the issue of missing or conflicting data include interpolation using spline surfaces \cite{Skytt2015}, kriging \cite{Reuter2007}, inverse distance weighting (IDW) \cite{Shepard:1968} and triangular irregular networks (TINs) \cite{Luedeling2007}.
These methods perform differently with respect to the type of terrain they are used to fill; smooth, sharp or containing irregular patterns.

In this letter we apply transfer learning techniques to train a model to be able to recover general features that are found in digital elevation/surface models.
In this way we avoid the need to apply different methods to different terrain types.
The need for human input is also limited to post-processing.

Our results are obtained by transferring to DEMs the recent successes of generative modeling techniques in the research field of \emph{image inpainting}, meaning the problem of filling missing regions of an image with data that appear plausible, in the sense of human interpretation. In the context of DEMs, plausibility of the filled data is not necessarily sufficient. In many cases, one would wish to accurately reproduce the missing part of the elevation model. However, in many cases this is an unrealistic goal due to various limitations. We therefore restrict our attention in this letter to providing a satisfactory fill for the various regions considered.


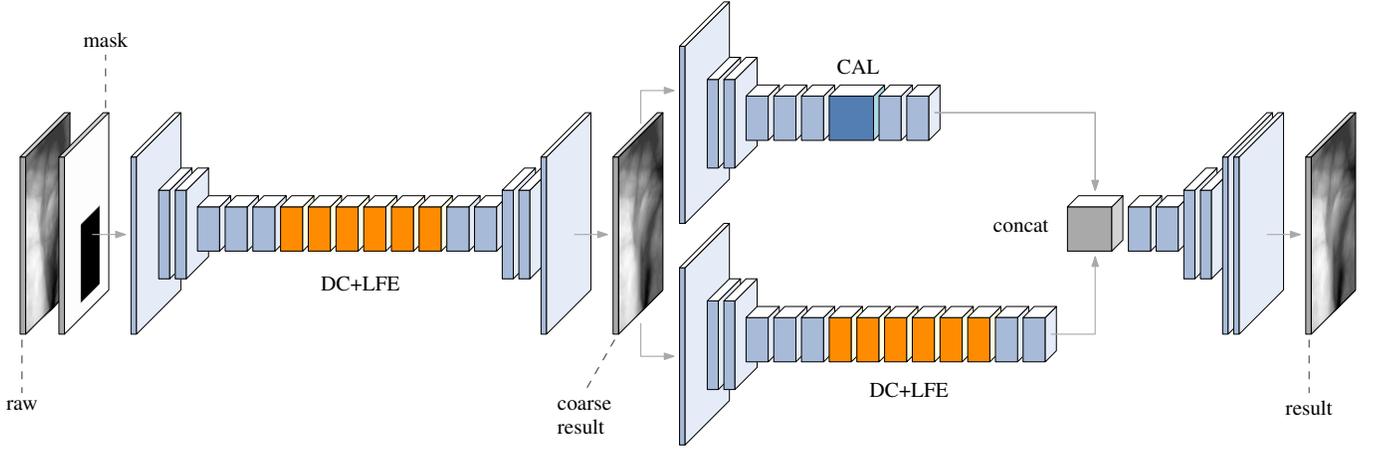
\begin{figure*}[!t]
\centering
\footnotesize
\begin{tikzpicture}
  \node[anchor=south west,inner sep=0] (image) at (0,0) {\includegraphics[width=.98\textwidth]{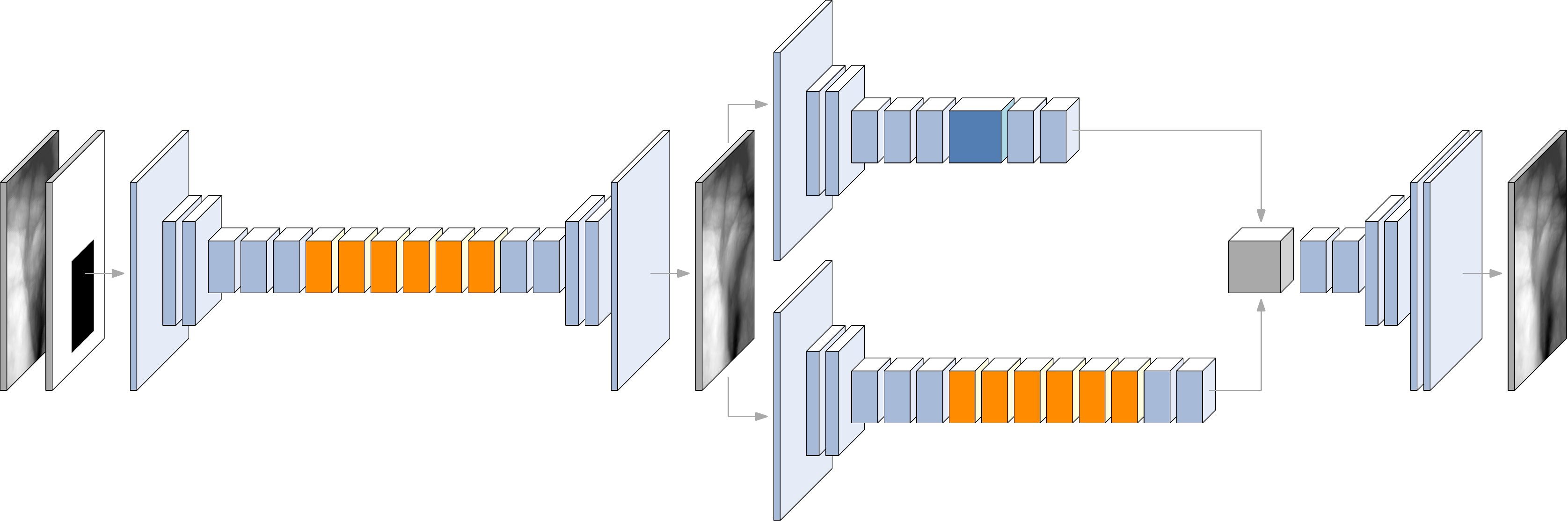}};
  \begin{scope}[x={(image.south east)},y={(image.north west)}]

  \node[below] at (0.255,0.4) {DC+LFE};
  \node[below] at (0.665,0.16) {DC+LFE};

  \node[above] at (0.627,0.82) {CAL};

  \node[left] at (0.775,0.495) {concat};

  \draw[dashed,-, darkgray](0.002,0.24) -- (0.002,0.12);
  \node[below] at (0.002,0.12) {raw};

  \draw[dashed,-, darkgray](0.0645,0.76) -- (0.0645,0.88);
  \node[above] at (0.0645,0.88) {mask};

  \draw[dashed,-, darkgray](0.445,0.24) -- (0.422,0.12);
  \node[below, text width=20pt] at (0.422,0.12) {coarse result};

  \draw[dashed,-, darkgray](0.964,0.24) -- (0.964,0.12);
  \node[below] at (0.964,0.12) {result};

  \end{scope}
\end{tikzpicture}

\caption{The generative void filling model utilized for missing DEM value completion. We highlight the dilated convolutions (DC), local feature extraction (LFE) modules, and the contextual attention layer (CAL).\label{fig:model}}
\end{figure*}


\section{Background in Generative Models}

\emph{Generative models} form a branch of unsupervised machine learning techniques that estimate the (joint) probability distribution underlying given data. For complex high-dimensional probability distributions, such as for DEM data generation, it is not feasible to learn this distribution explicitly. It is however possible to obtain an implicit description through a model capable of generating samples from an estimated distribution.

\emph{Generative adversarial networks} (GANs) \cite{Goodfellow2014} form a highly promising framework for training a model that generates such samples. One reason for this is that the adversarial loss in GANs is known to catch highly recognizable salient features not picked up by mean squared error (MSE) \cite{LotterKC15}. At the core of a GANs are two adversaries attempting to outwit one another: A \emph{generator} learns to generate fake samples that are supposed to look real, and a \emph{discriminator} learns to distinguish real data from fake samples. When these adversaries are carefully balanced, both become better over training time.

Initially GANs were difficult to train due to this balancing act. This situation is remedied to some extent by the recently proposed \emph{Wasserstein GAN} (WGAN) \cite{Arjovsky2017}, which capitalizes on better theoretical properties of the Wasserstein-1 distance --- also known as Earth Mover's (EM) distance, as it measures the effort necessary for moving the estimated probability density to the true density.

Although GANs at their inception showed great results for small images, this success was initially difficult to scale up to high-resolution data. Constraining the network architecture to only convolutional layers, the \emph{Deep Convolutional GAN} (DCGAN) \cite{Radford2015} brought simplicity, deeper models, flexibility in image resolution, and insight into what GANs learn. Adding dilated convolutions \cite{Yu2015MultiScaleCA} takes this one step further, bringing back an enhanced receptive field previously handled by fully connected layers.

These techniques form the foundation for a recent wave of deep generative inpainting networks \cite{Iizuka2017,Li2017,Pathak2016}. Further improvements to training stability and speed are obtained by adding to the loss function a spatially discounted reconstruction loss, as well as local and global critics to ensure local consistency and global coherence, and a contextual attention mechanism to capture relevant features at a distance \cite{Yu2018}.

\section{Methodology}

\subsection{Problem formulation and notation}

We will consider preprocessed digital elevation/surface models in GeoTIFF format, in which the data forms a grid with a single height value for every position $(i,j)$.
Let $\Dw=(d_p)\in\mathbb{R}^{m\times n}$ be a partial digital elevation model, where $p$ is an abbreviation for pixel referring to the coordinates $(i,j)$ of a point on the DEM grid and $d_p$ is the corresponding height value. Partial means that some pixel values are considered void. A binary matrix $\Mw = (m_p)\in \{0,1\}^{m\times n}$ acts as a mask representing the void regions of $\Dw$. We refer to pixels $p$ for which $m_p = 0$ as \textit{known}, and \textit{unknown} otherwise.

\begin{problem}\label{problem}
Given an initial partial elevation model $\Dw^0$ and the corresponding mask $\Mw$, construct a complete elevation model $\Dw$ with semantically plausible generated values for the masked regions.
\end{problem}


\begin{algorithm}[H]
\caption{DEM Void Filling}
\begin{algorithmic}[1]
\renewcommand{\algorithmicrequire}{\textbf{Input:}}
\renewcommand{\algorithmicensure}{\textbf{Output:}}

\Require \emph{partial DEM $\Dw^0=(d^0_p)$, mask $\Mw=(m_p)$
\Statex[1] blending weight function $\beta$, blending width $w$
\Statex[1] paraboloid fitting window radius $r$}

\Ensure \emph{predicted DEM $\Dw$}

\State $\Dw \leftarrow G(\Dw^0, \Mw)$ \Comment{\emph{initial result}}

\State partition unknown pixels to sets $R_k$ of pixels
\Statex with $L_1$-distance of $k$ from a known pixel

\For {$k\leftarrow 1, \ldots, w$}

  \For {$p=(i,j)\in R_k$}

    \State compute the set $X$ of known pixels of $\Dw^0$
    \Statex[2] in subgrid $[{i-r},{i+r}]\times[{j-r},{j+r}]$

    \State perform least squares paraboloid fitting to X
    \Statex[2] $f^*\leftarrow \argmin\limits_f \sum\limits_{q \in X}\left[f(q)-d_{q}\right]^2$

    \State $d^0_p \leftarrow f^*(p)$\Comment \emph{approximate $C^2$ extension of $\Dw^0$}
    \State $\alpha\leftarrow \beta(\frac{k-1}{w})$ \Comment \emph{blending weight}
    \State $d_p \leftarrow \left(1-\alpha\right)d^0_p + \alpha d_p$ \Comment \emph{blend}
  \EndFor
  \State $m_{R_k} \leftarrow 0$ \Comment \emph{label pixels in $R_k$ as known}
\EndFor
\end{algorithmic}
\label{algorithm}
\end{algorithm}


\subsection{Main algorithm}

Our method solves Problem \ref{problem} in two steps. Initially, we get a complete elevation model $\Dw$ by employing a deep generative network $G$, which has been trained to complete missing data values while respecting relevant features of the surrounding regions. The second step involves optional post-processing of the result of Step 1 for obtaining a smooth transition between the initial known region and the prediction provided by $G$.

Algorithm \ref{algorithm} is a complete description of the proposed solution which we will analyze in detail in the sequel.

\subsection{Model Architecture}

The proposed DEM void filling generative model $G$ is an adaptation of the generative image inpainting model presented in \cite{Yu2018}; see Figure \ref{fig:model}. This model demonstrates promising results for texture-like images, which we leverage to transferring topographic patterns in our setting.

The input consists of the two arrays $\Dw^0$ and $\Mw$. We focus on size $256\times 256$ arrays for our implementation.
The training set is generated by artificially removing randomly generated rectangular regions from the ground truth provided by complete GeoTIFFs from the Norwegian Mapping Authority. This data source was chosen for two reasons. First of all it is openly available, facilitating reproducible research. Secondly, we hypothesize that the variety in Norwegian topography makes it well-suited for generalization to other regions of the world.

Following the coarse-to-fine network approach of \cite{Yu2018}, the missing region is at first filled with a coarse prediction, which is fed to a second network for refinement, before the end result $\Dw = G(\Dw^0,\Mw)$.
The coarse prediction stage is a dilated deep convolutional encoder-decoder network trained with reconstruction loss, generating a smooth initial guess for the contents of the hole.
The refinement stage contains two parallel encoders, one implementing the contextual attention mechanism and the other a dilated deep convolutional encoder, merged as input to a single decoder generating a prediction for the completed grid.

For improved local feature aggregation---necessary for remote sensing applications---both dilation components of the model utilize the recent \emph{Local Feature Extraction} (LFE) module \cite{Hamaguchi2018}, which consists of convolutional layers masks of size $3\times 3$ and with first increasing and then decreasing dilations 2-4-8-8-4-2.

The composed network is trained end-to-end with $\ell_1$ reconstruction loss, global and local Wasserstein GAN Gradient-Penalty (WGAN-GP) adversarial loss \cite{gulrajani+al-2017-wasserstein-arxiv}.
The reconstruction loss is spatially discounted, in the sense that hallucination is stronger the further away one is from known data. For further network specifics, hyperparameters, and examples, see \cite{DEM-fill}.


\begin{figure}[!t]
\centering
\footnotesize
\begin{tikzpicture}
  \node[anchor=south west,inner sep=0] (image) at (0,0) {\includegraphics[cframe=lightgray 6pt, width=0.485\linewidth]{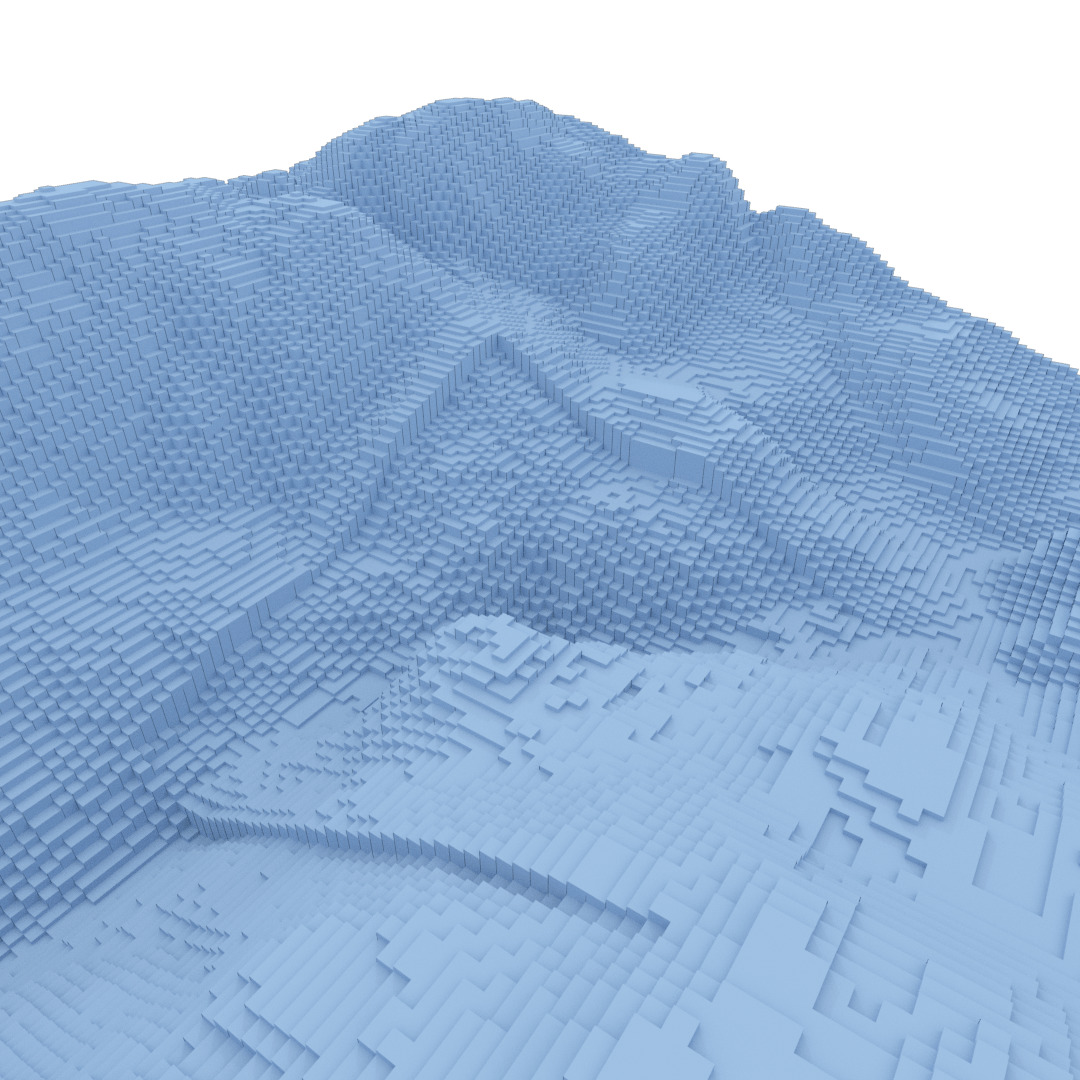}};
  \begin{scope}[x={(image.south east)},y={(image.north west)}]


  \end{scope}
\end{tikzpicture}
\hfill
\begin{tikzpicture}
  \node[anchor=south west,inner sep=0] (image) at (0,0) {\includegraphics[cframe=lightgray 6pt, width=0.485\linewidth]{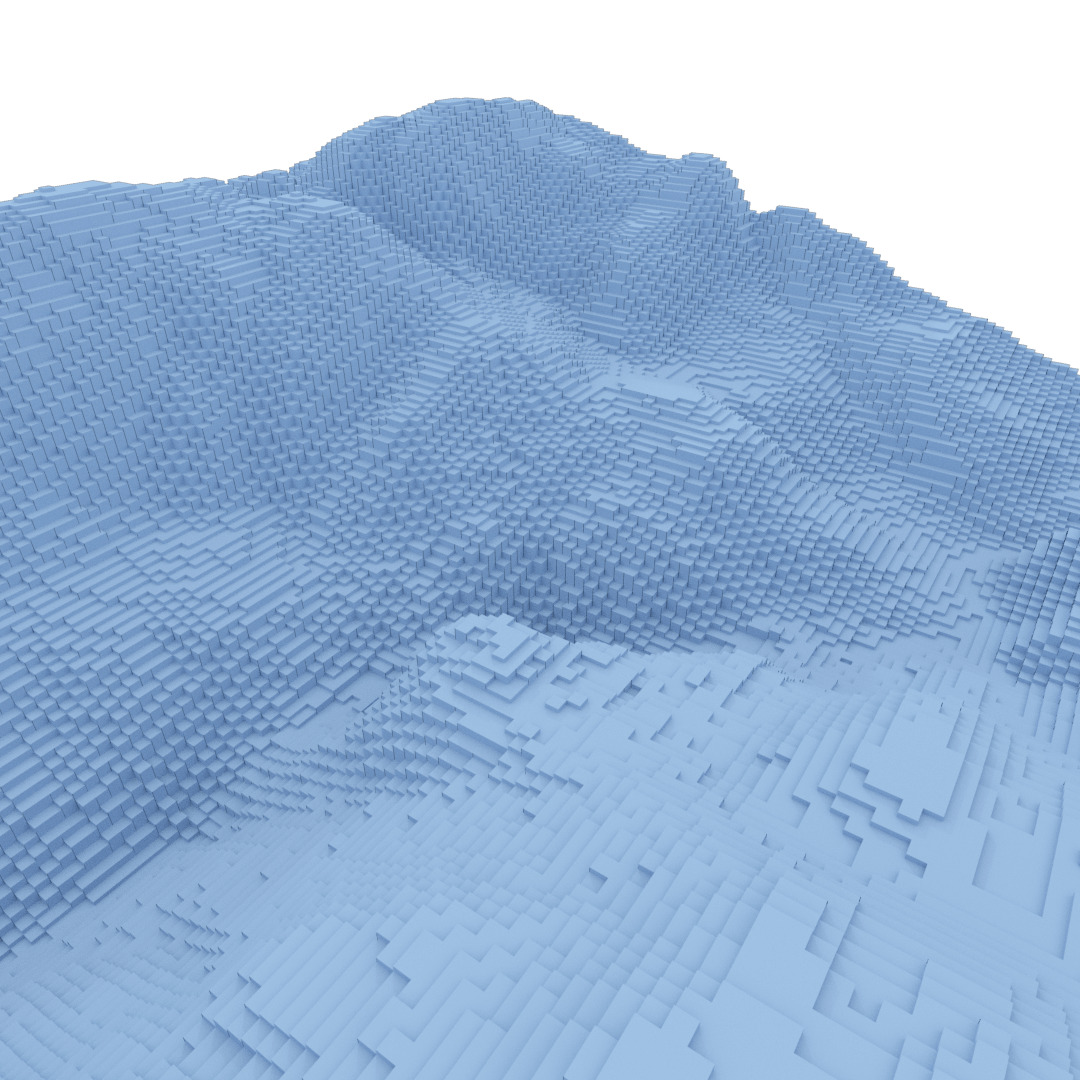}};
  \begin{scope}[x={(image.south east)},y={(image.north west)}]


  \end{scope}
\end{tikzpicture}

\caption{Detail from a landscape DEM with exaggerated boundary discontinuity (left) and the result of the post-processing boundary blending step (right).\label{fig:boundary}}
\end{figure}


\subsection{Boundary post-processing}

We propose an optional post-processing step to remedy any boundary artifacts between the known edges and the generated elevation values. The intuition behind the procedure is that we compute the approximate $C^2$-continuous extension of the known region boundary and blend it appropriately with the resulting DEM. Figure \ref{fig:boundary} demonstrates an example of the post-processing step.

We partition the unknown pixels to sets $R_k$, each containing pixels with $L_1$-distance of $k$ from a known pixel.
For $k=1,\ldots,w$, where $w\in\mathbb{N}$ is the chosen width of the blending region, we update the current boundary $R_k$ with the following procedure.
For each boundary pixel $p=(p_i, p_j)\in R_k$, let $X$ be the set of known pixels in $[p_i-r,p_i+r]\times[p_j-r,p_j+r]$, i.e., the known entries of the submatrix of size $(2r+1)\times (2r+1)$ centered at $p$. The value of $r$ is chosen by the user (we use 3 or 4). The best fitting paraboloid
$$
f(i,j) = Ai^2 + Bij + Cj^2 + Di + Ej + F
$$
in the least squares sense is given by the solution
$$
f^* = \argmin\limits_f\sum\limits_{q \in X}\left[f(q)-d_{q}\right]^2
$$
and approximates the local curvature of the DEM. We assign $f^*(p)$ to the value of $p$, and repeat the process for all pixels of $R_k$. The boundary $R_k$ is labeled as known and we move to the next boundary $R_{k+1}$. The entire extension procedure is repeated $w$ times to achieve an extension of width $w$.

The $C^2$ extension is then blended with the predicted result in the following manner. We choose a strictly increasing bijective blending function $\beta: [0,1]\rightarrow [0,1]$.
For our experiments, we use a sigmoid blending function.
The final value $d^f_p$ at pixel $p\in R_k$, $k=1,\ldots,w$ is the linear blend of the value $d^0_p$ of the $C^2$-extended initial elevation model and the value $d_p$ of the result from the generative network $G$, that is
$$
d^f_p = (1-\alpha_k)d^0_p+\alpha_k d_p,
$$
where $\alpha_k = \beta(\frac{k-1}{w})$ is the blending weight.

\section{Experiments and Results}\label{sec:results}

We trained two separate models for our experiments. Model $G_1$ was trained on rectangular $\si{2\metre}$-resolution DEMs of the three largest cities in Norway, namely Oslo, Trondheim, and Bergen, while model $G_2$ was trained on $\si{10\metre}$-resolution DEM of Western and Eastern Norway.


We compare our generators $G_1$, $G_2$ with two traditional methods to void filling; inverse distance weighting interpolation (IDW) and splines.
For a fair comparison, no post-processing was used.
The implementation of IDW is taken from GDAL \cite{GDAL} with the option of two $3\times 3$ smoothing filter passes. For the spline approach we utilize locally refined (LR) B-splines \cite{Dokken2013,Skytt2015}.
This letter is too short to contain a complete description of LR B-splines, but for the purpose of our application we expect them to perform at least as well as tensor-product B-splines \cite{Reuter2007}.

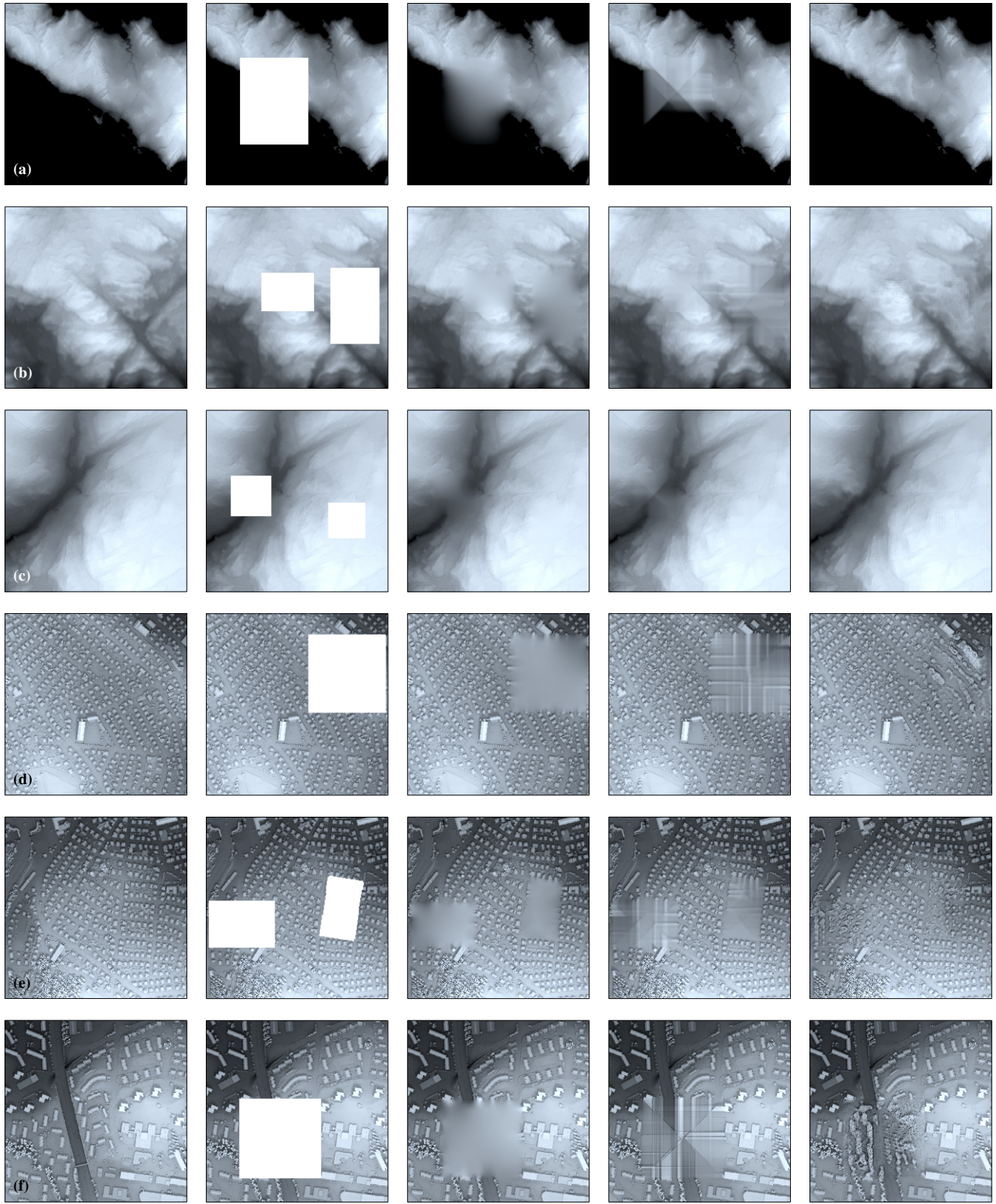
\begin{figure*}
\centering
\setlength{\extrarowheight}{92pt}
\begin{tabularx}{\textwidth}{XXXXc}

\begin{tikzpicture}
  \node[anchor=south west,inner sep=0] (image) at (0,0) {\includegraphics[frame, width = 0.18\textwidth]{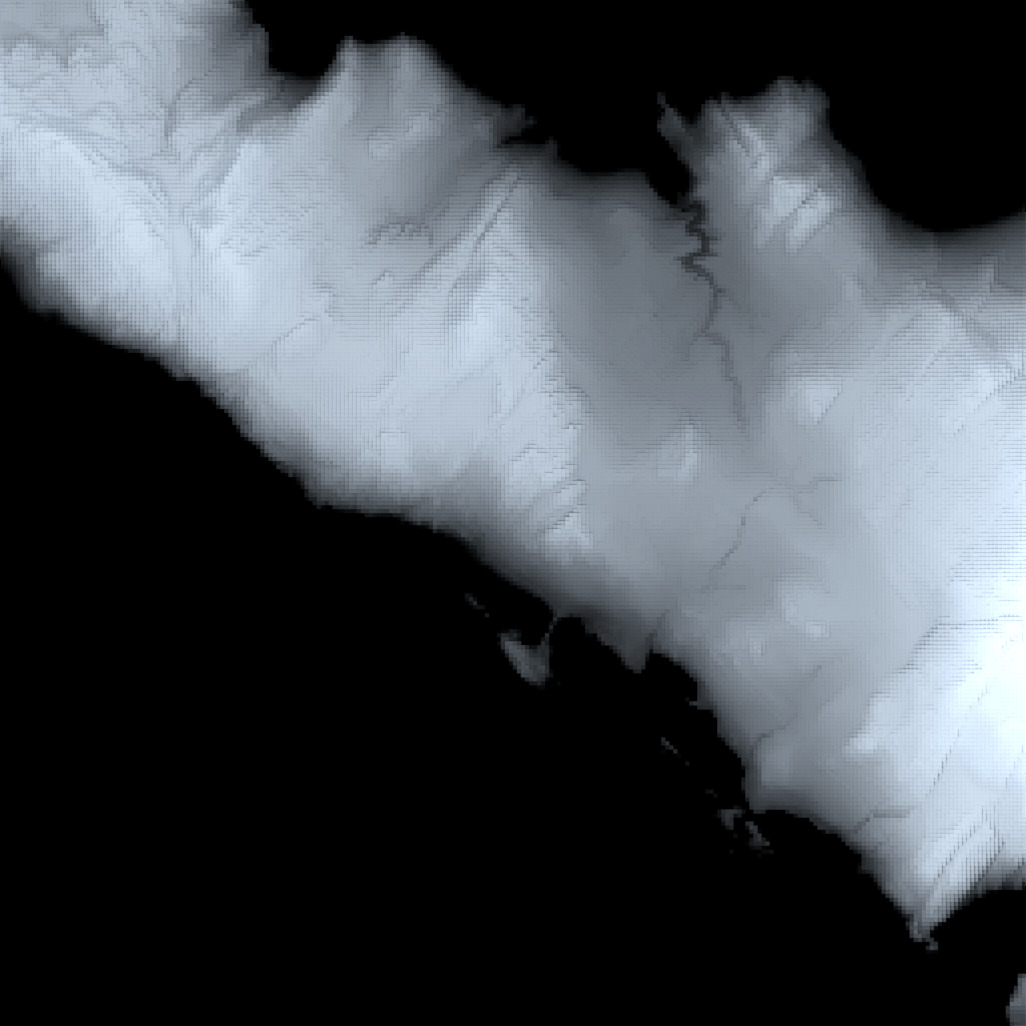}};
  \begin{scope}[x={(image.south east)},y={(image.north west)}]

  \node[above right] at (0.01,0.01) {\color{lightgray!10}\footnotesize\textbf{(a)}};

  \end{scope}
\end{tikzpicture} &
\includegraphics[frame, width = 0.18\textwidth]{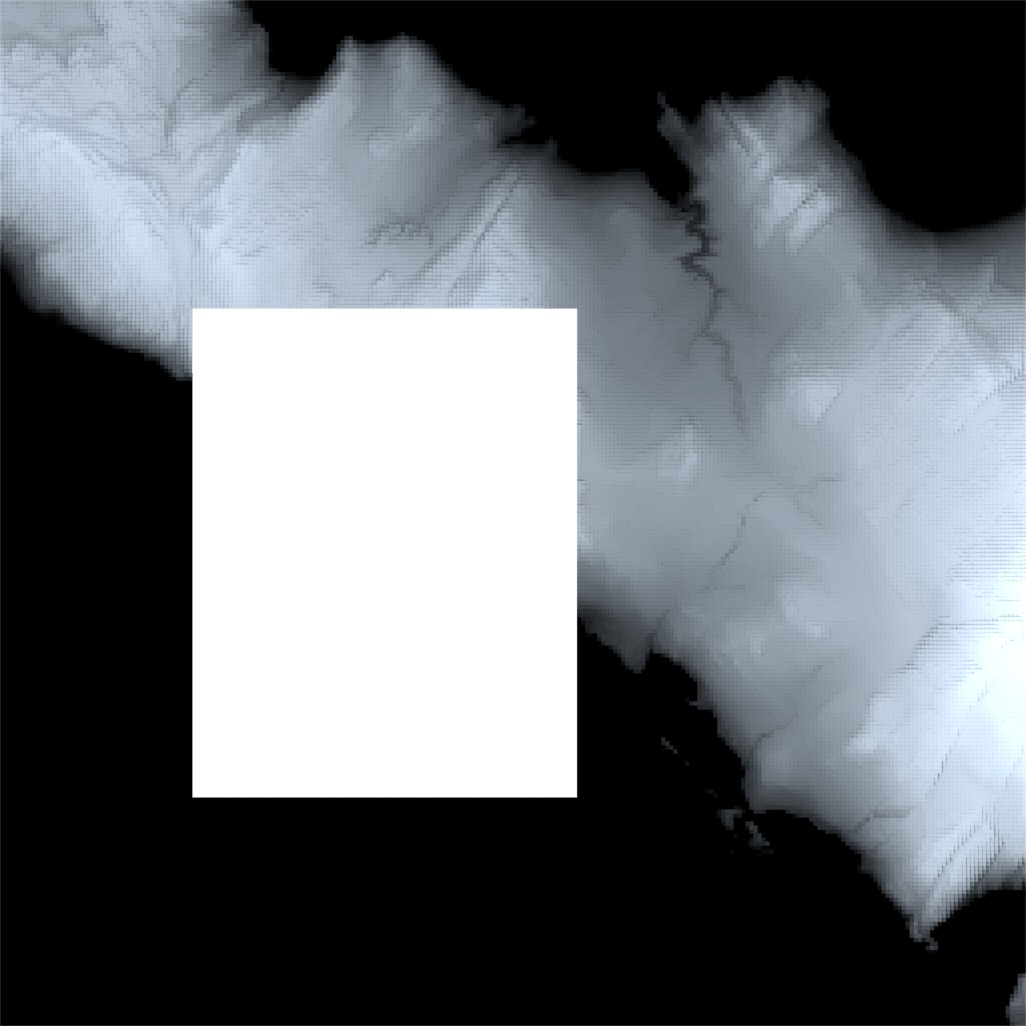} &
\includegraphics[frame, width = 0.18\textwidth]{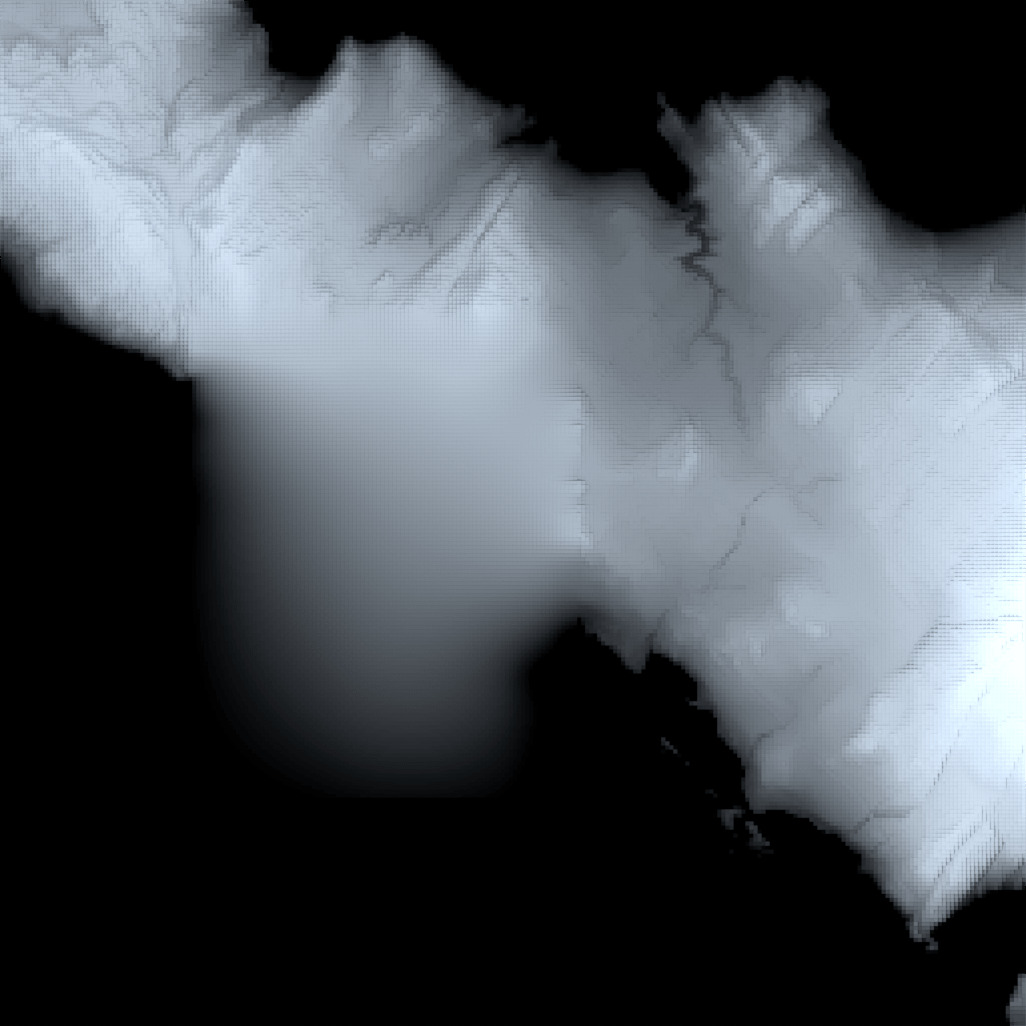} &
\includegraphics[frame, width = 0.18\textwidth]{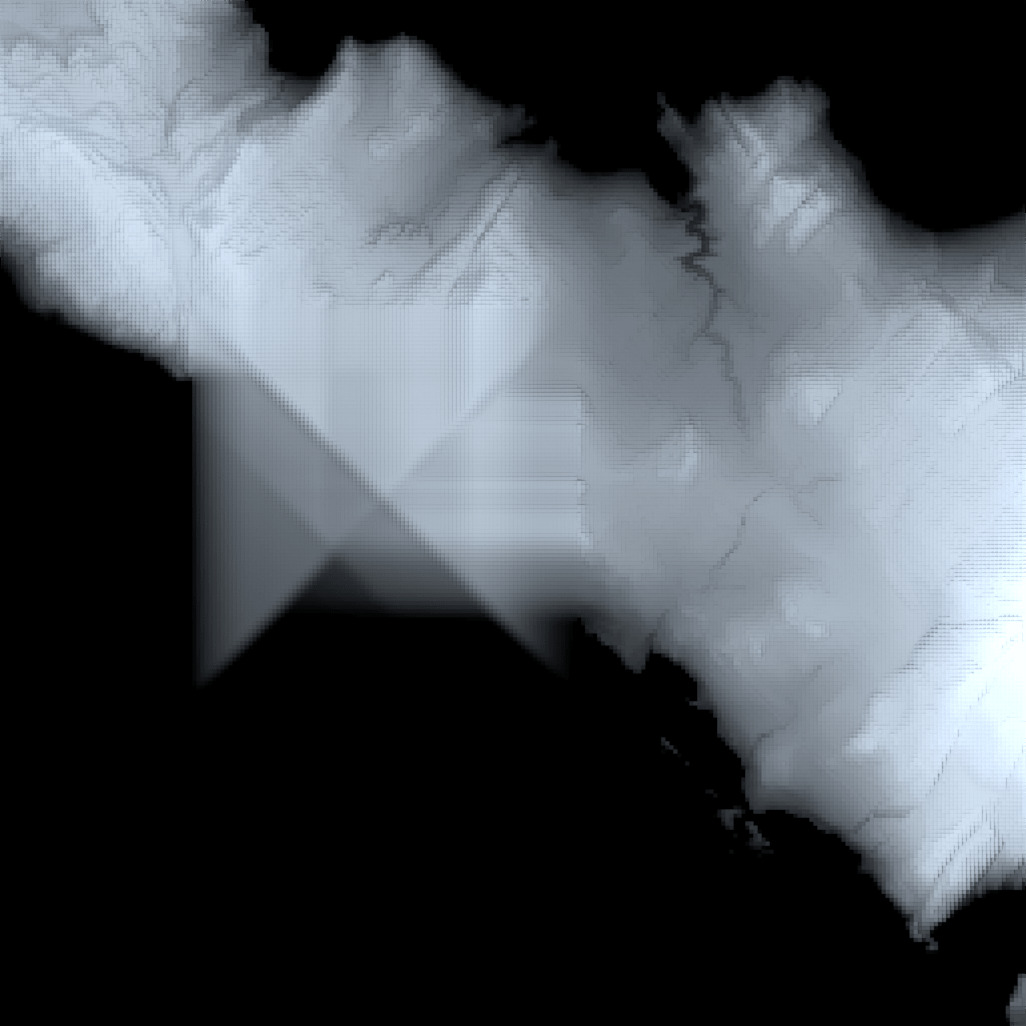} &
\includegraphics[frame, width = 0.18\textwidth]{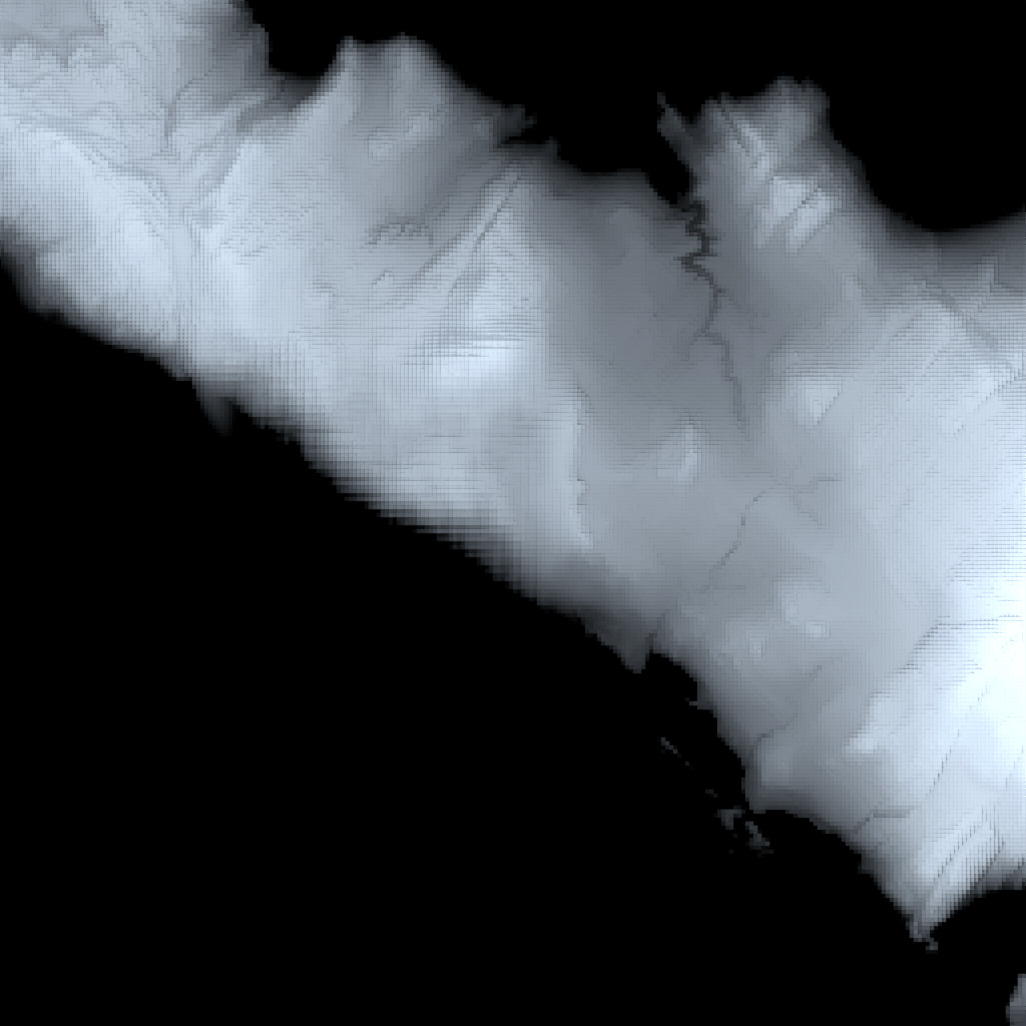}\\

\begin{tikzpicture}
  \node[anchor=south west,inner sep=0] (image) at (0,0) {\includegraphics[frame, width = 0.18\textwidth]{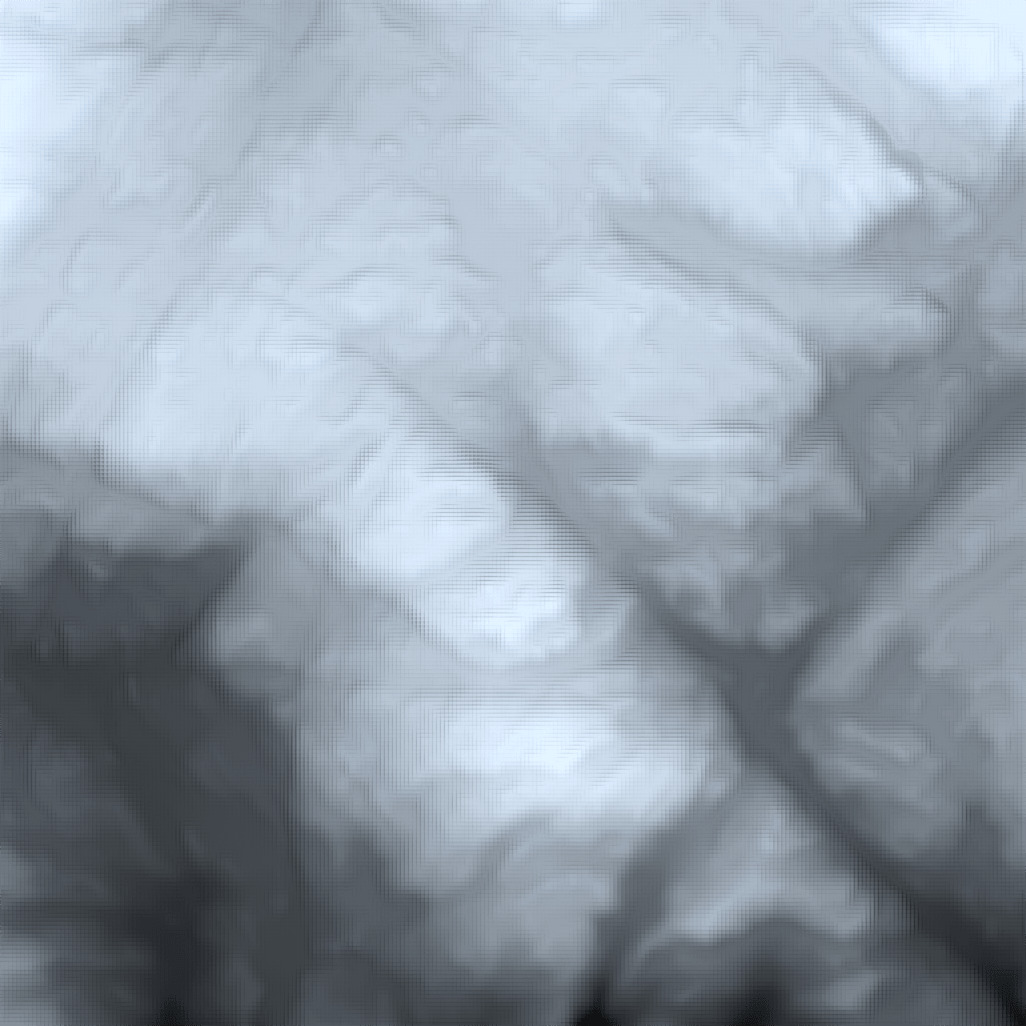}};
  \begin{scope}[x={(image.south east)},y={(image.north west)}]

  \node[above right] at (0.01,0.01) {\color{lightgray!10}\footnotesize\textbf{(b)}};

  \end{scope}
\end{tikzpicture} &
\includegraphics[frame, width = 0.18\textwidth]{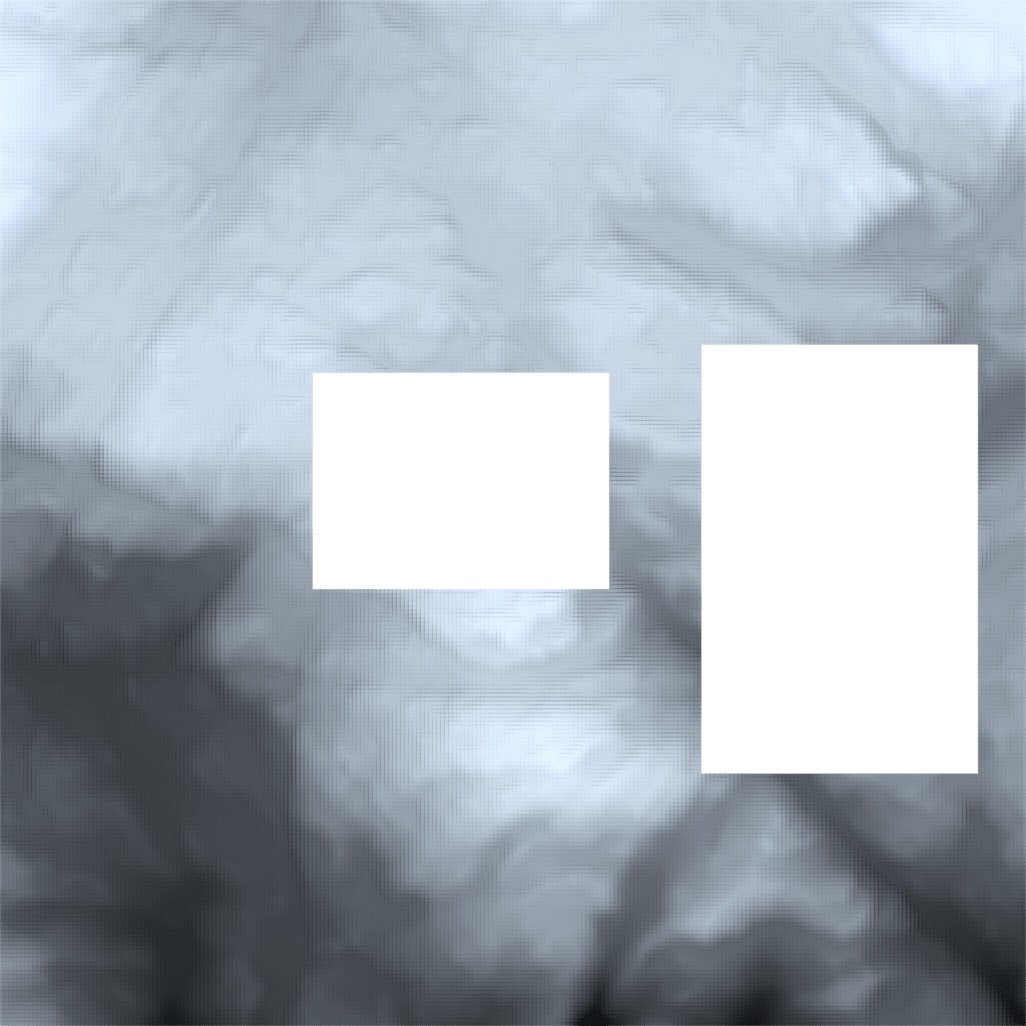} &
\includegraphics[frame, width = 0.18\textwidth]{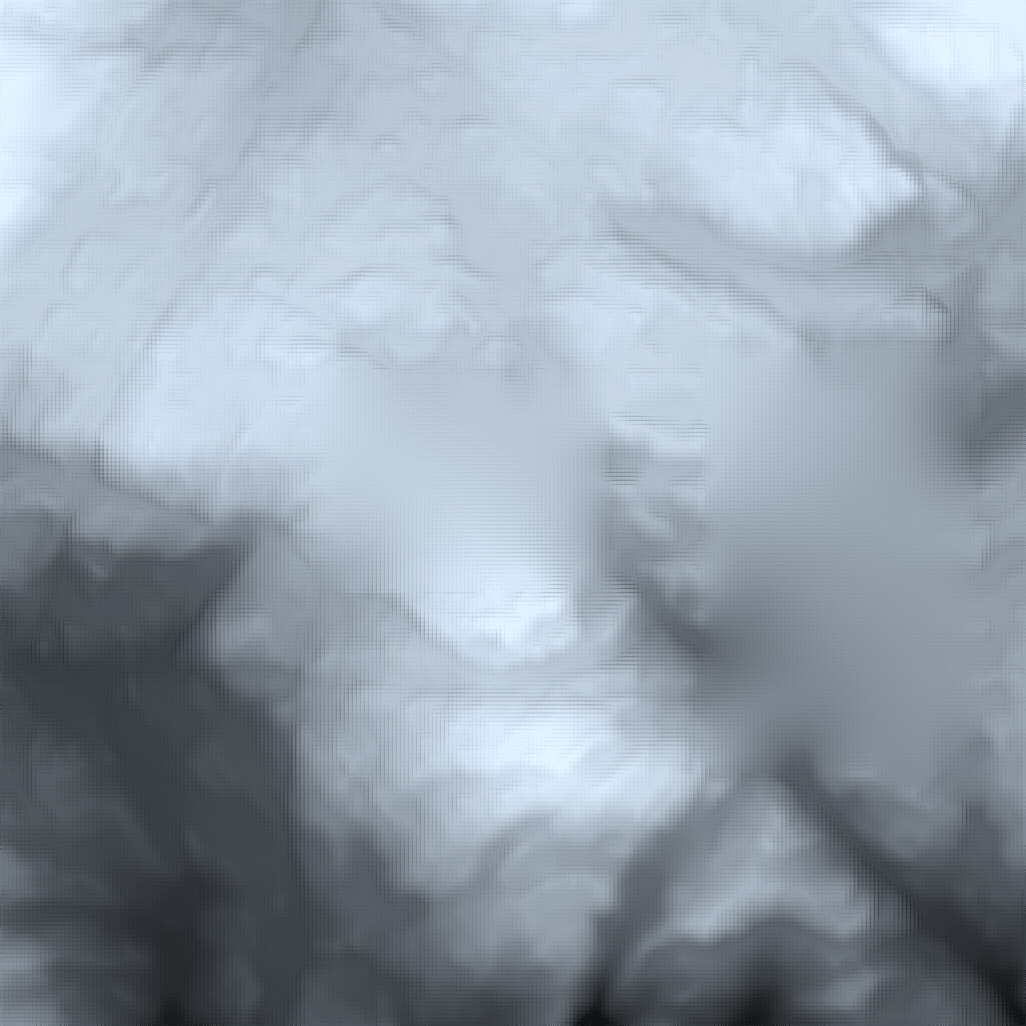} &
\includegraphics[frame, width = 0.18\textwidth]{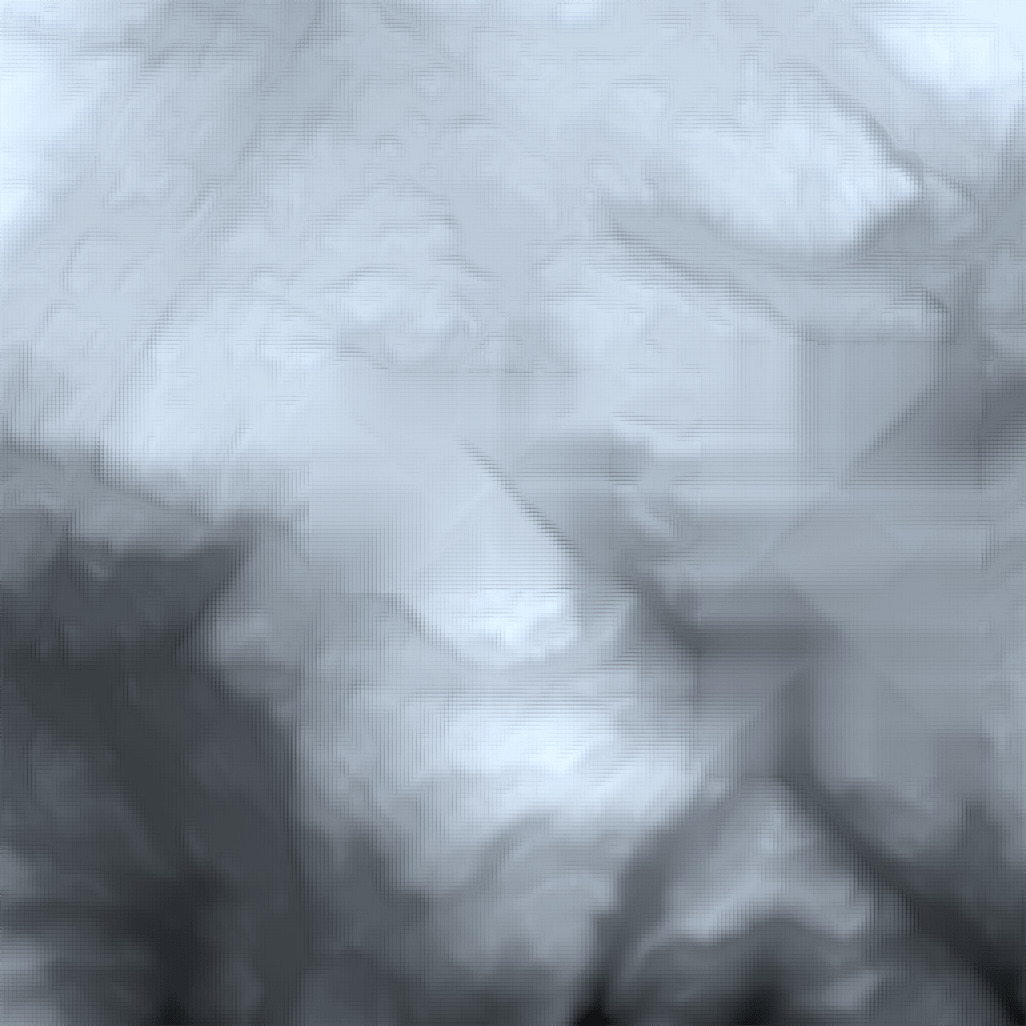} &
\includegraphics[frame, width = 0.18\textwidth]{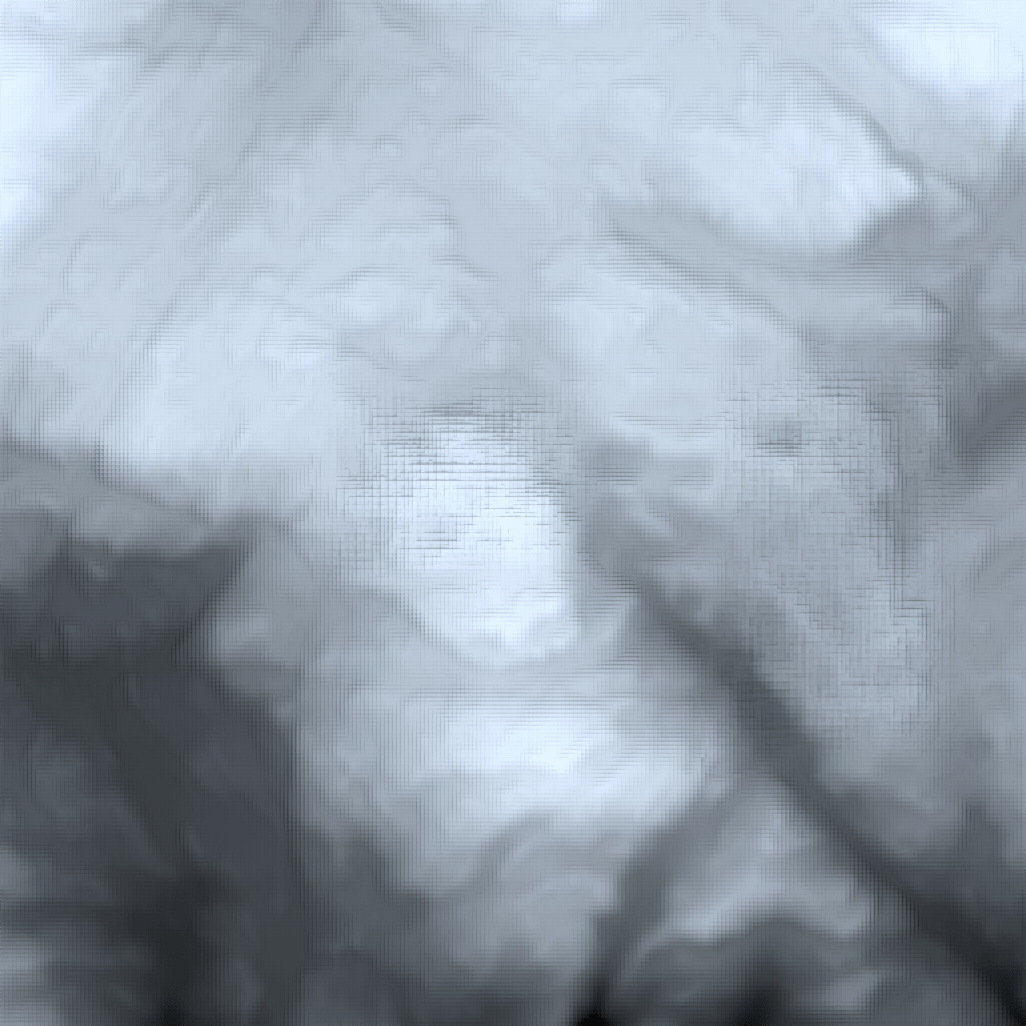}\\

\begin{tikzpicture}
  \node[anchor=south west,inner sep=0] (image) at (0,0) {\includegraphics[frame, width = 0.18\textwidth]{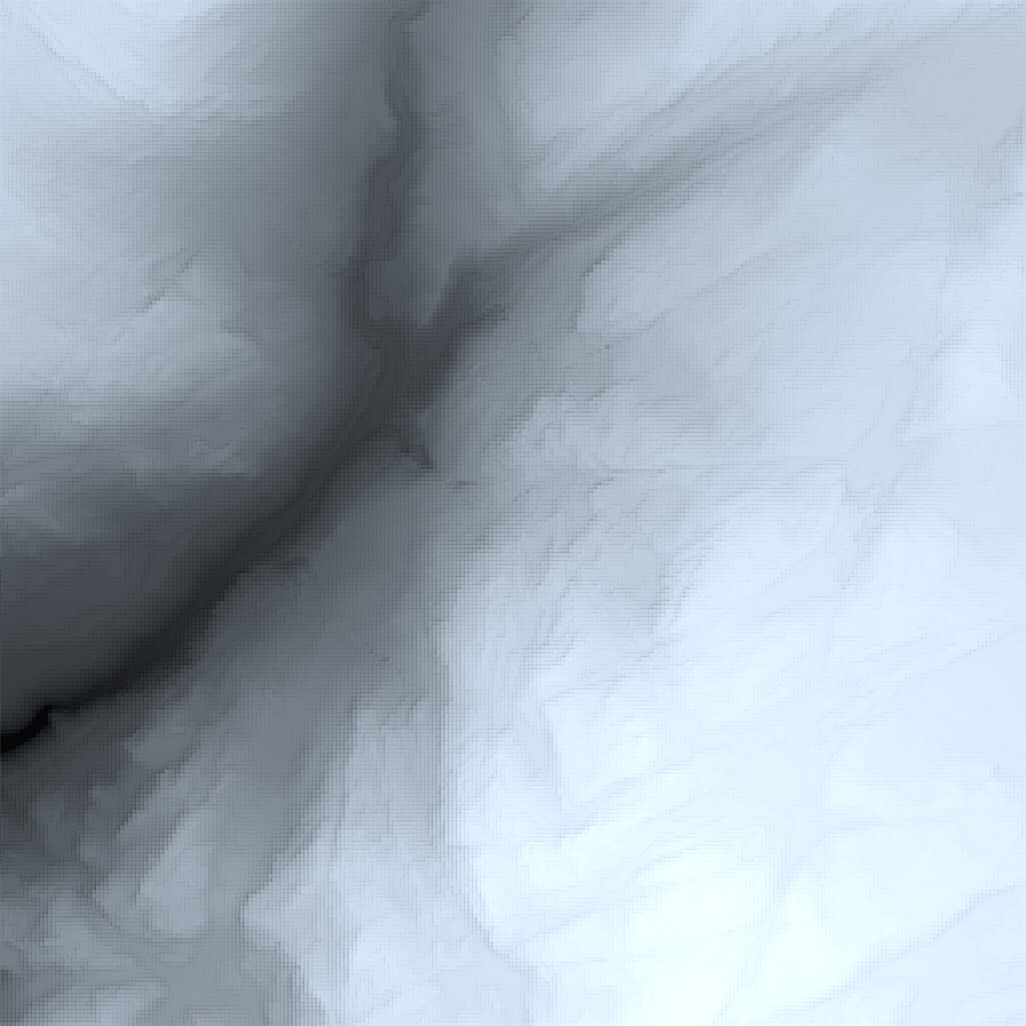}};
  \begin{scope}[x={(image.south east)},y={(image.north west)}]

  \node[above right] at (0.01,0.01) {\color{lightgray!10}\footnotesize\textbf{(c)}};

  \end{scope}
\end{tikzpicture} &
\includegraphics[frame, width = 0.18\textwidth]{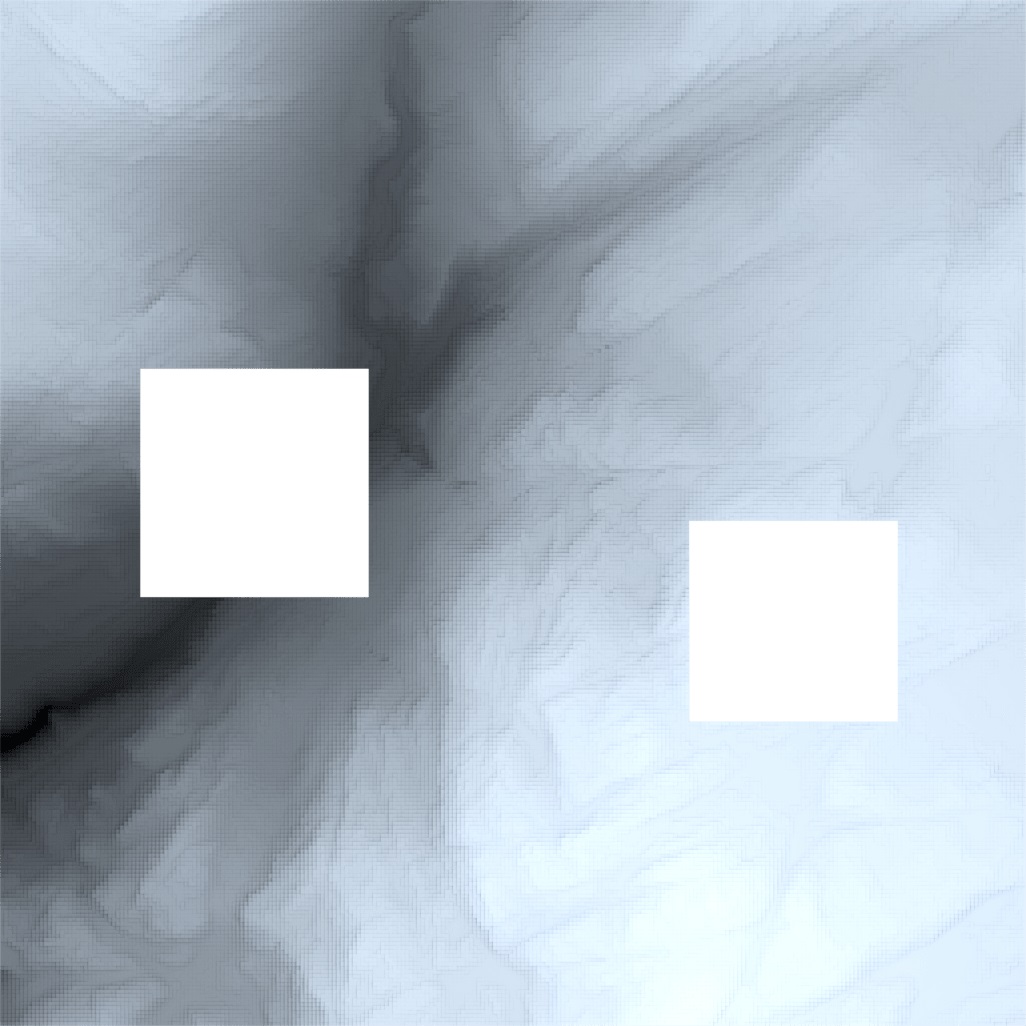} &
\includegraphics[frame, width = 0.18\textwidth]{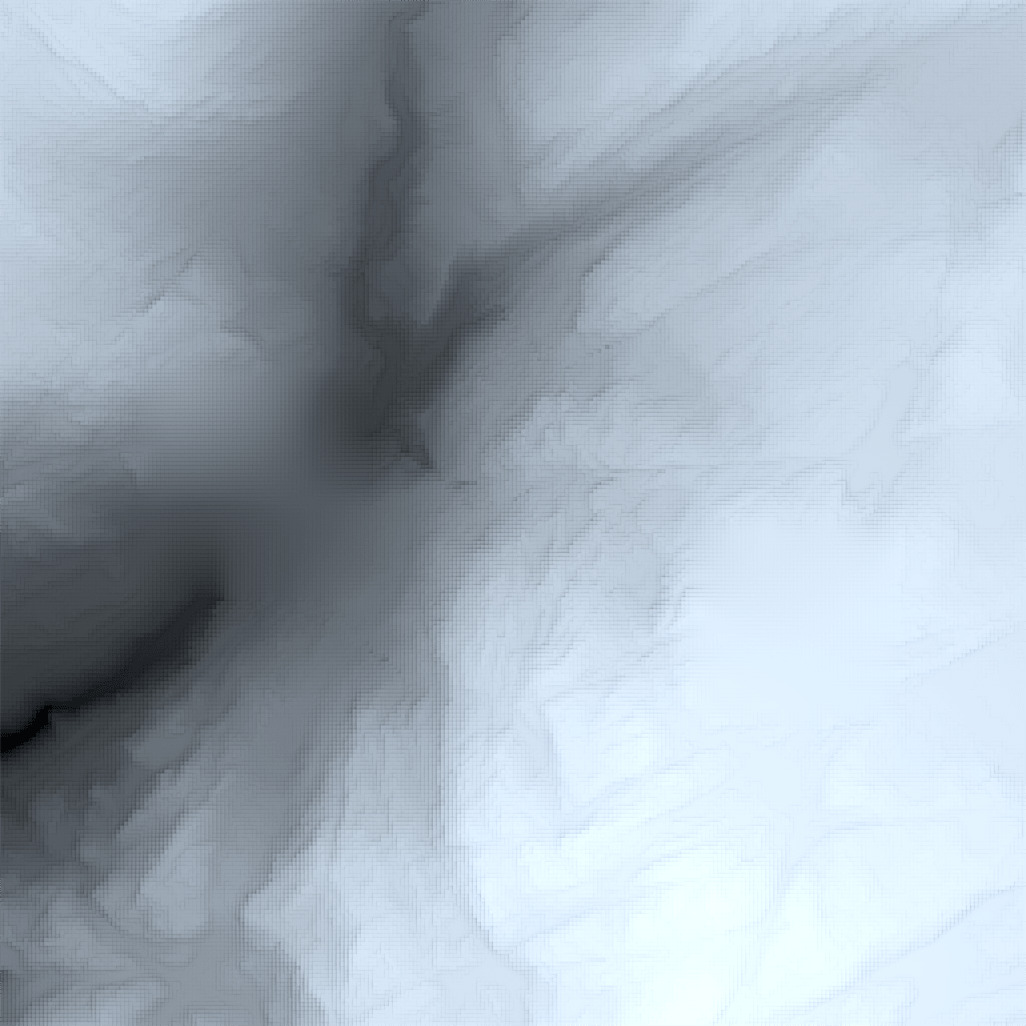} &
\includegraphics[frame, width = 0.18\textwidth]{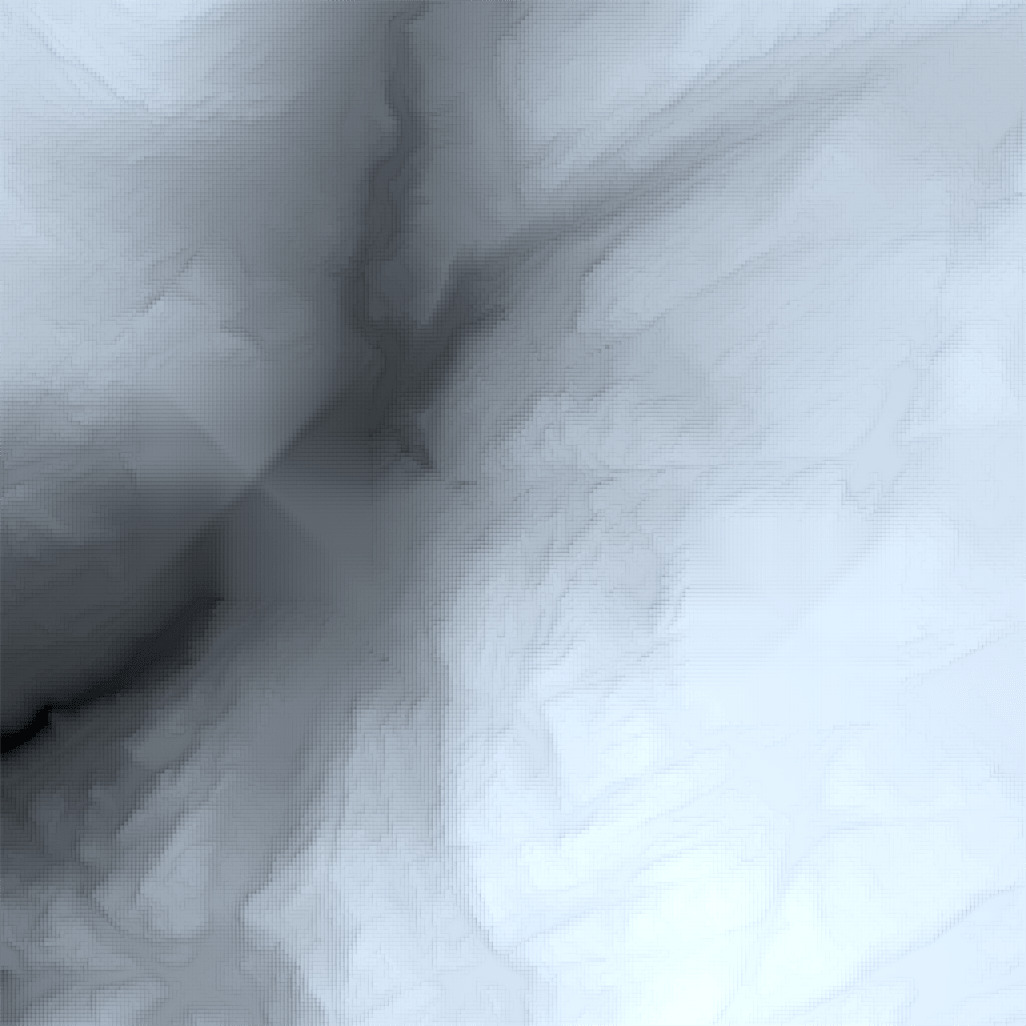} &
\includegraphics[frame, width = 0.18\textwidth]{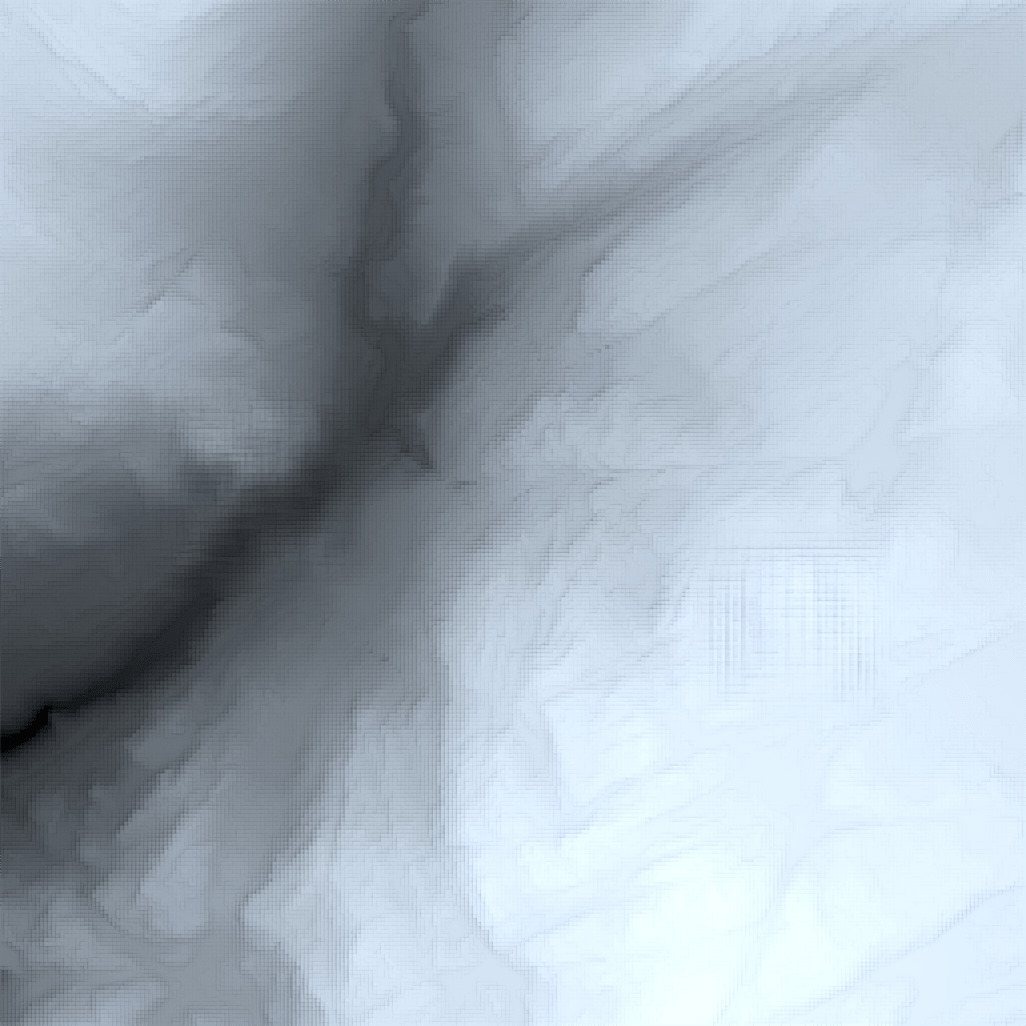}\\

\begin{tikzpicture}
  \node[anchor=south west,inner sep=0] (image) at (0,0) {\includegraphics[frame, width = 0.18\textwidth]{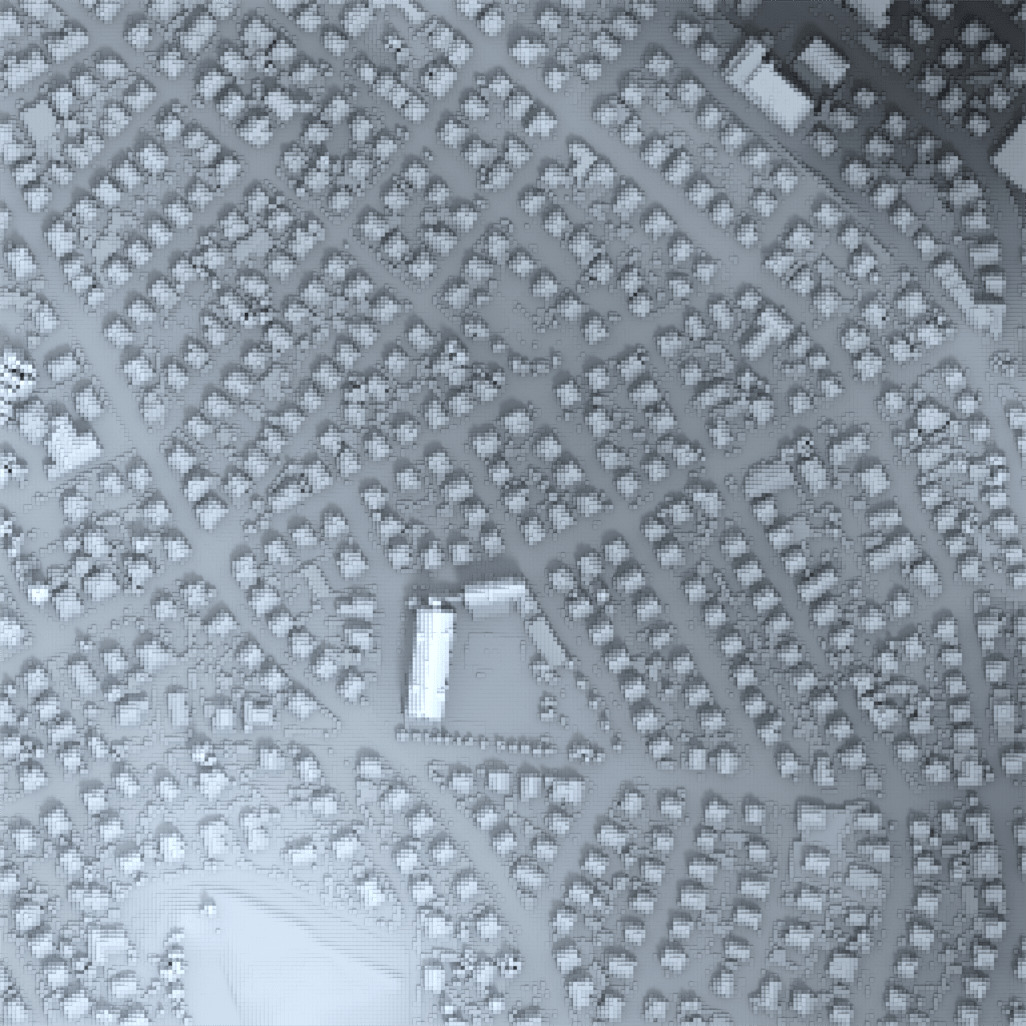}};
  \begin{scope}[x={(image.south east)},y={(image.north west)}]

  \node[above right] at (0.01,0.01) {\footnotesize\textbf{(d)}};

  \end{scope}
\end{tikzpicture} &
\includegraphics[frame, width = 0.18\textwidth]{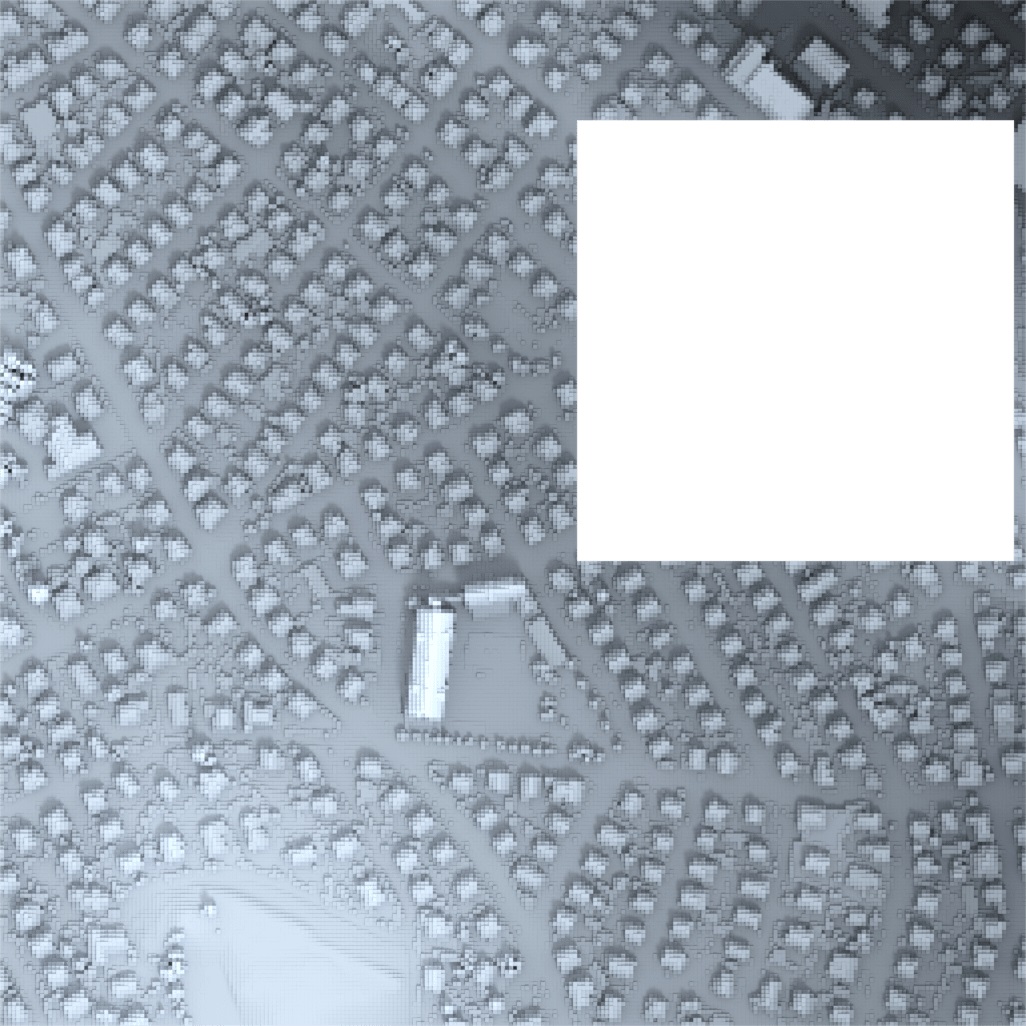} &
\includegraphics[frame, width = 0.18\textwidth]{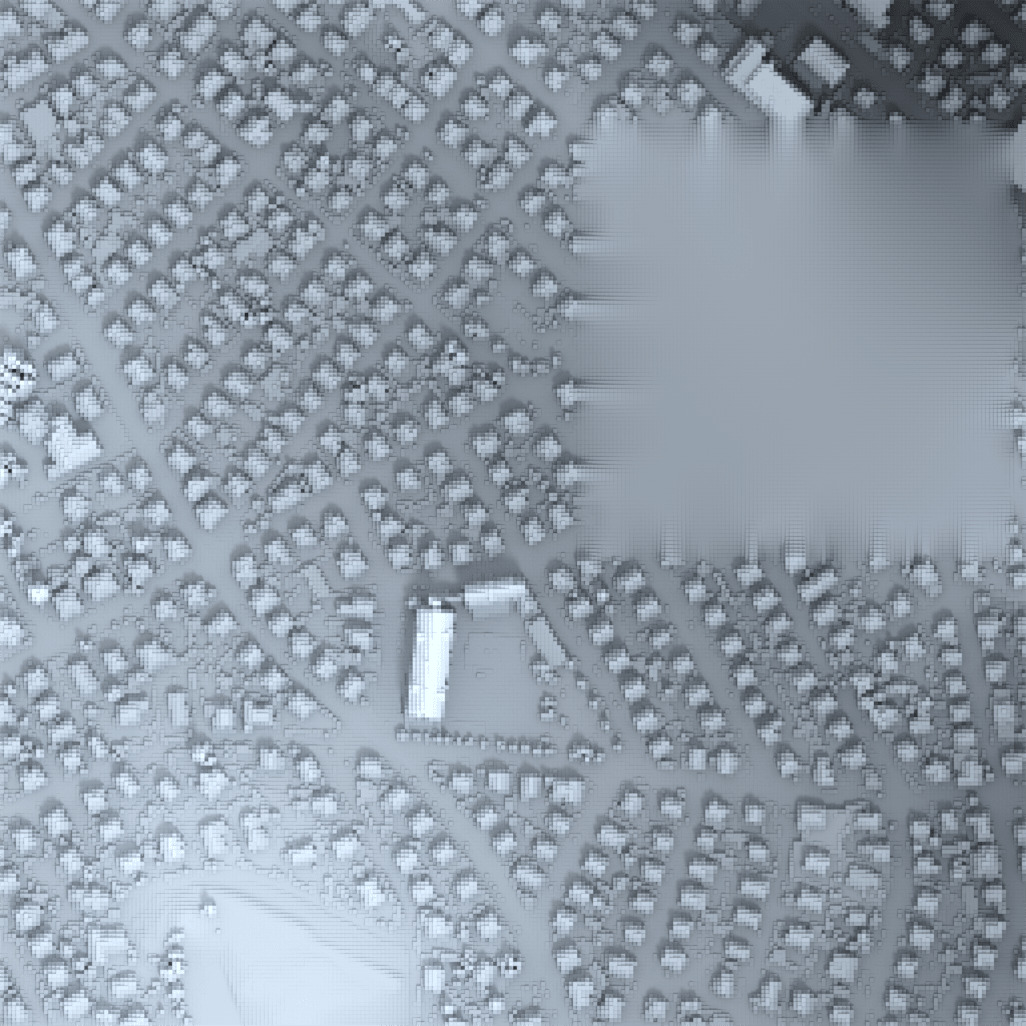} &
\includegraphics[frame, width = 0.18\textwidth]{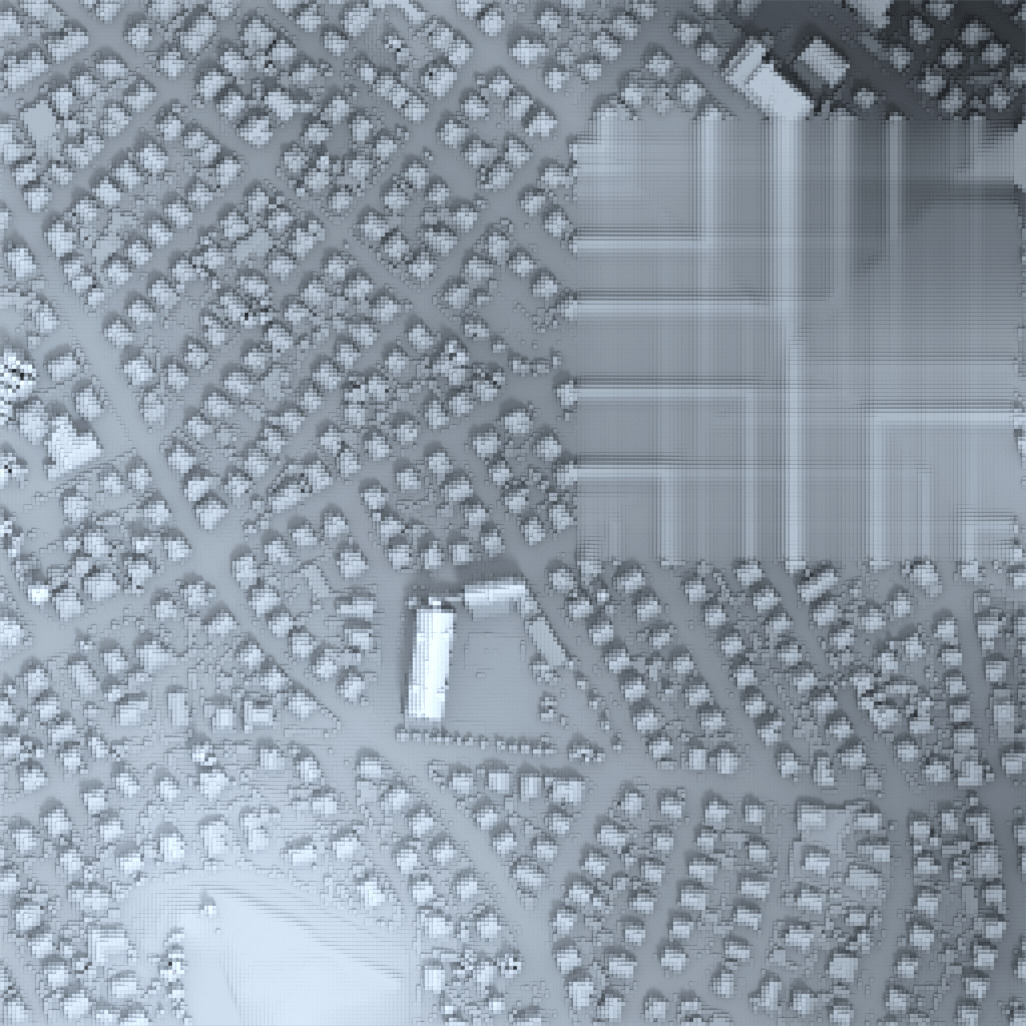} &
\includegraphics[frame, width = 0.18\textwidth]{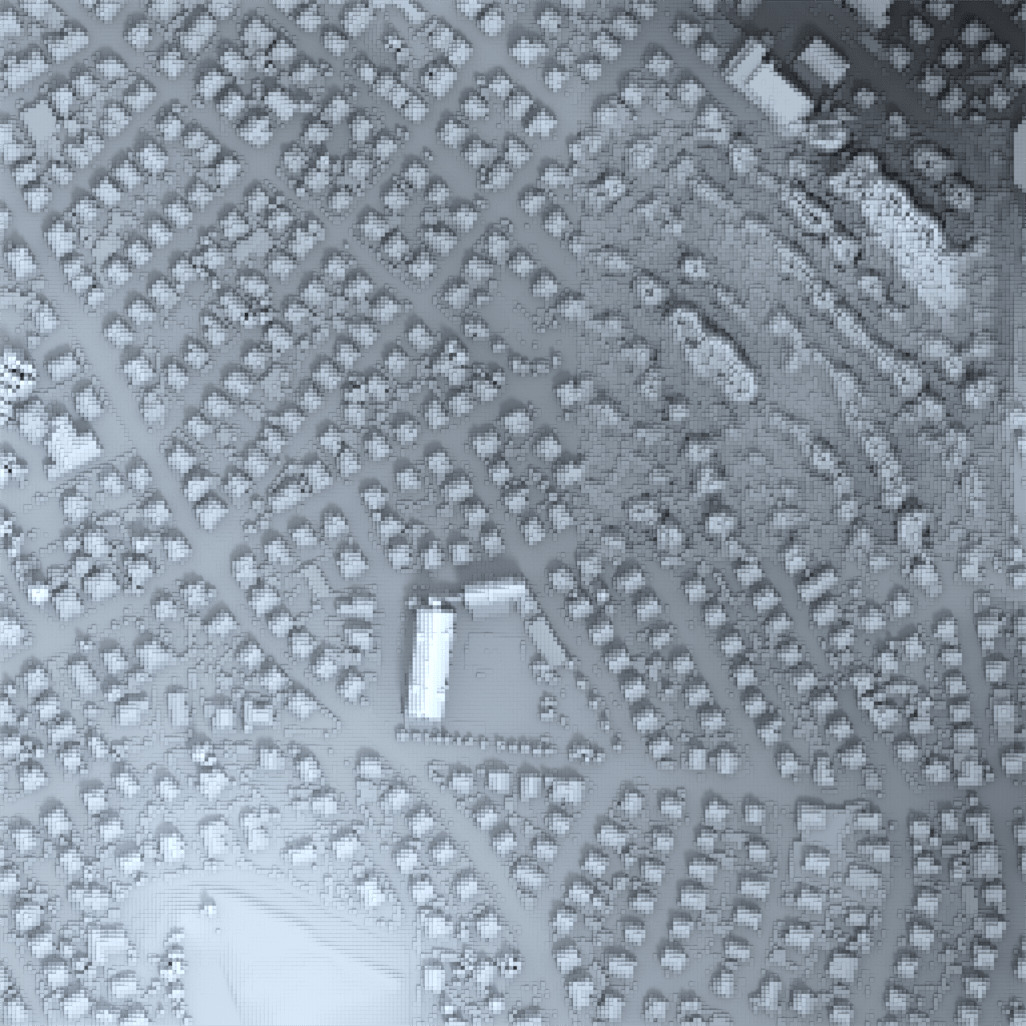}\\

\begin{tikzpicture}
  \node[anchor=south west,inner sep=0] (image) at (0,0) {\includegraphics[frame, width = 0.18\textwidth]{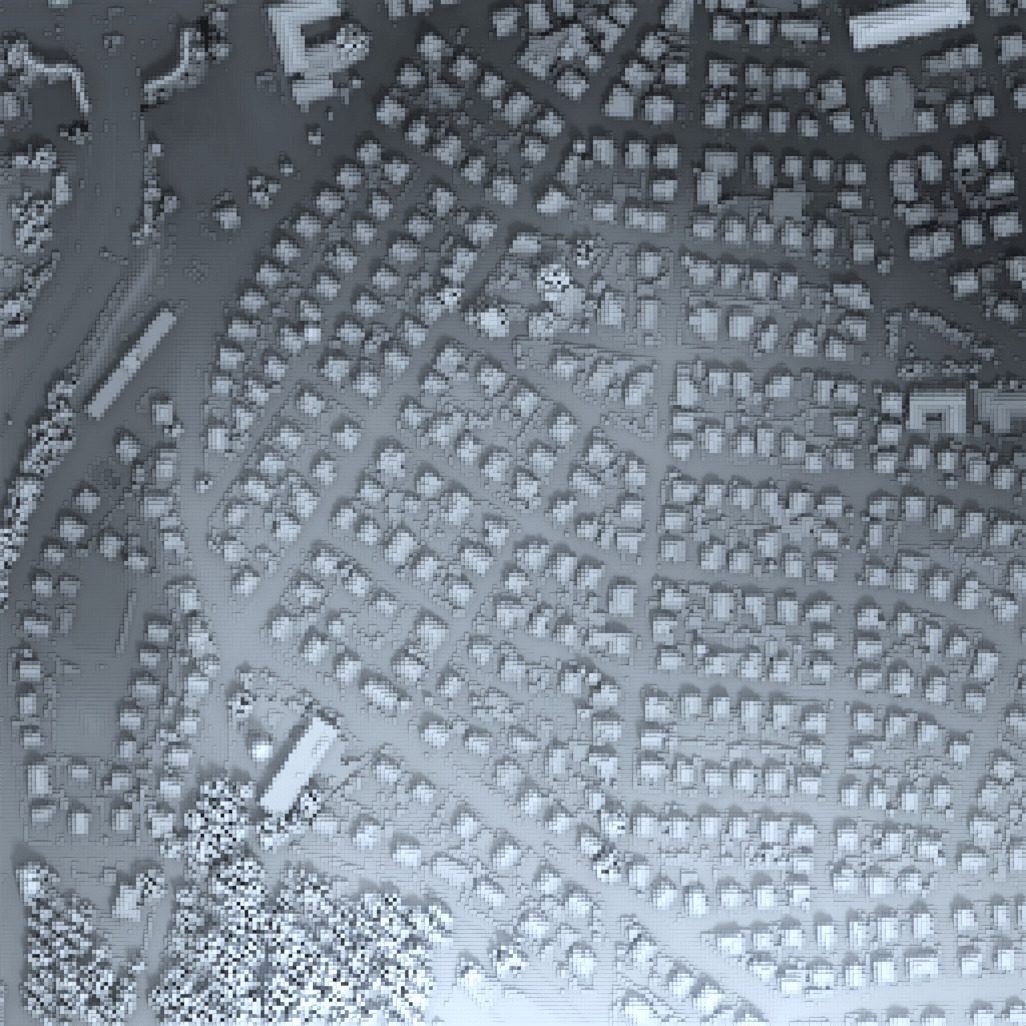}};
  \begin{scope}[x={(image.south east)},y={(image.north west)}]

  \node[above right] at (0.01,0.01) {\footnotesize\textbf{(e)}};

  \end{scope}
\end{tikzpicture} &
\includegraphics[frame, width = 0.18\textwidth]{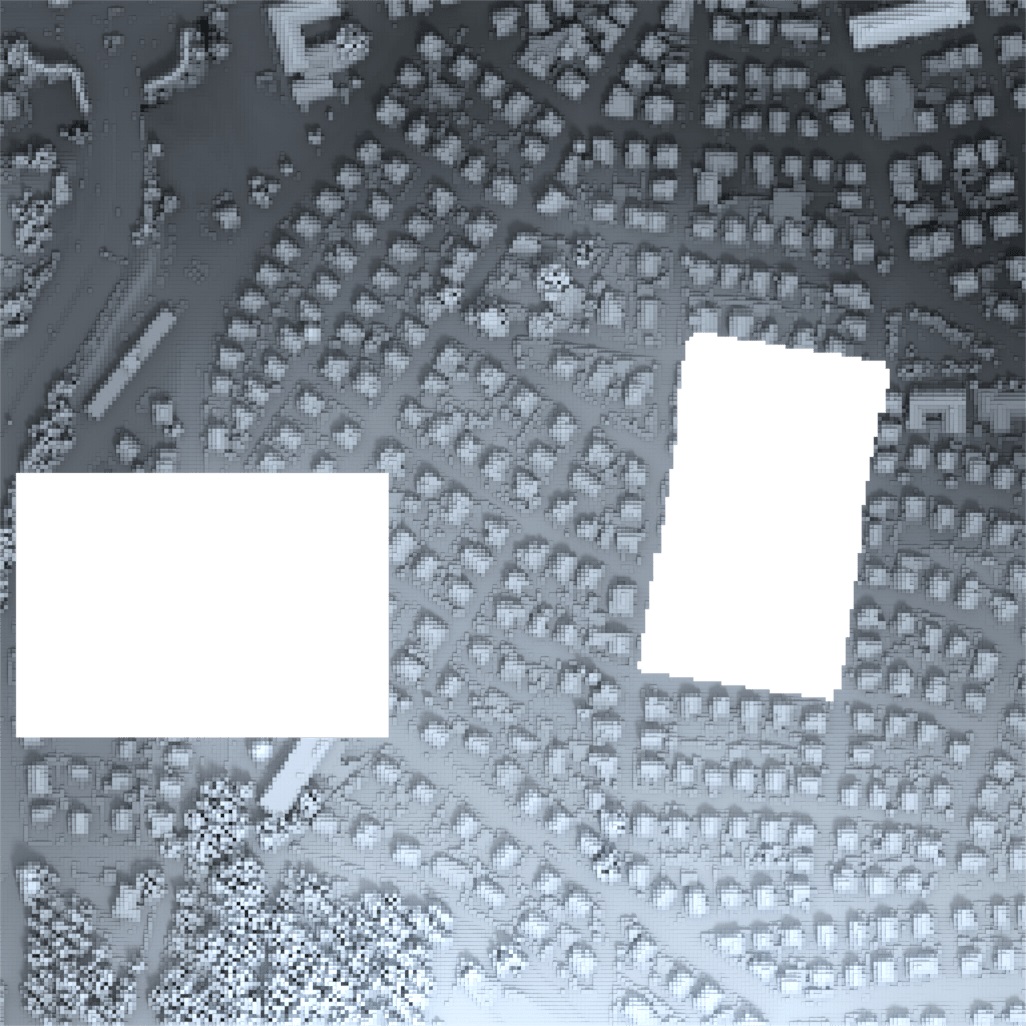} &
\includegraphics[frame, width = 0.18\textwidth]{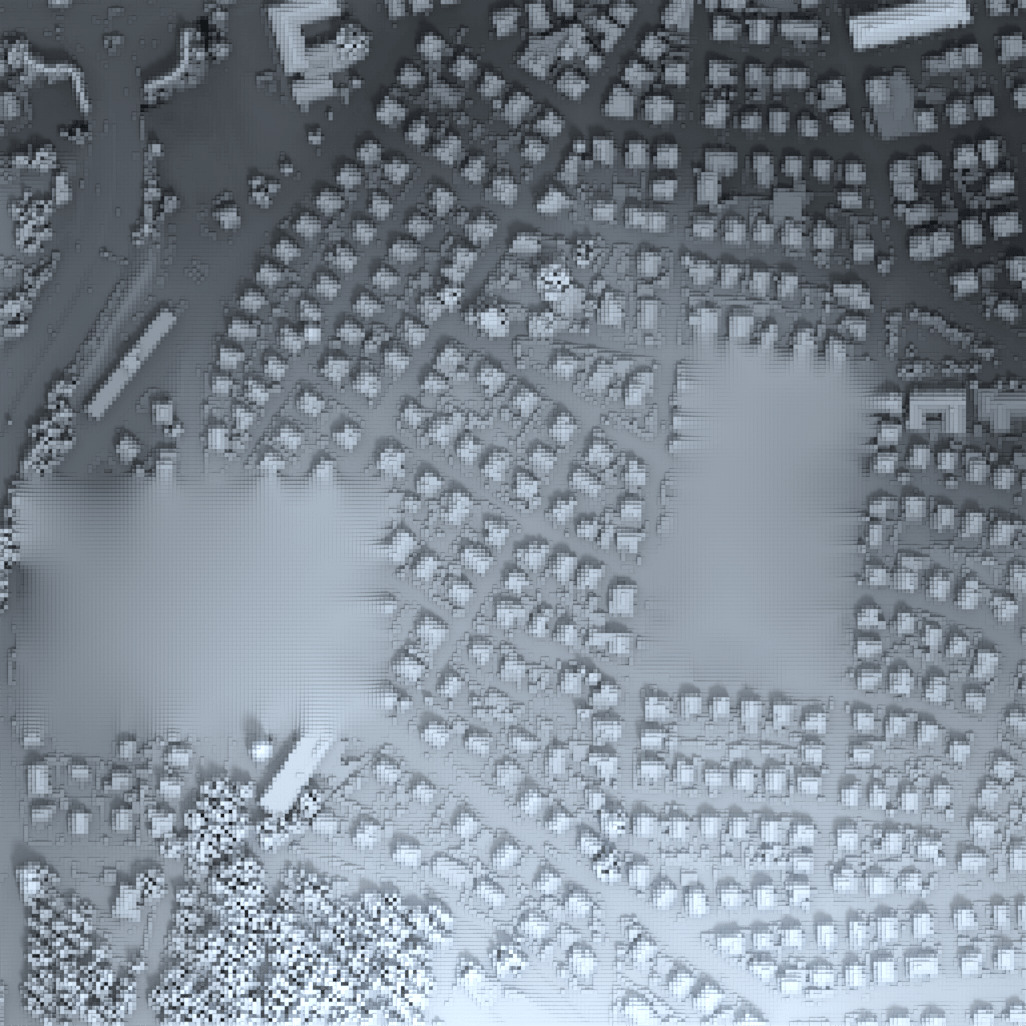} &
\includegraphics[frame, width = 0.18\textwidth]{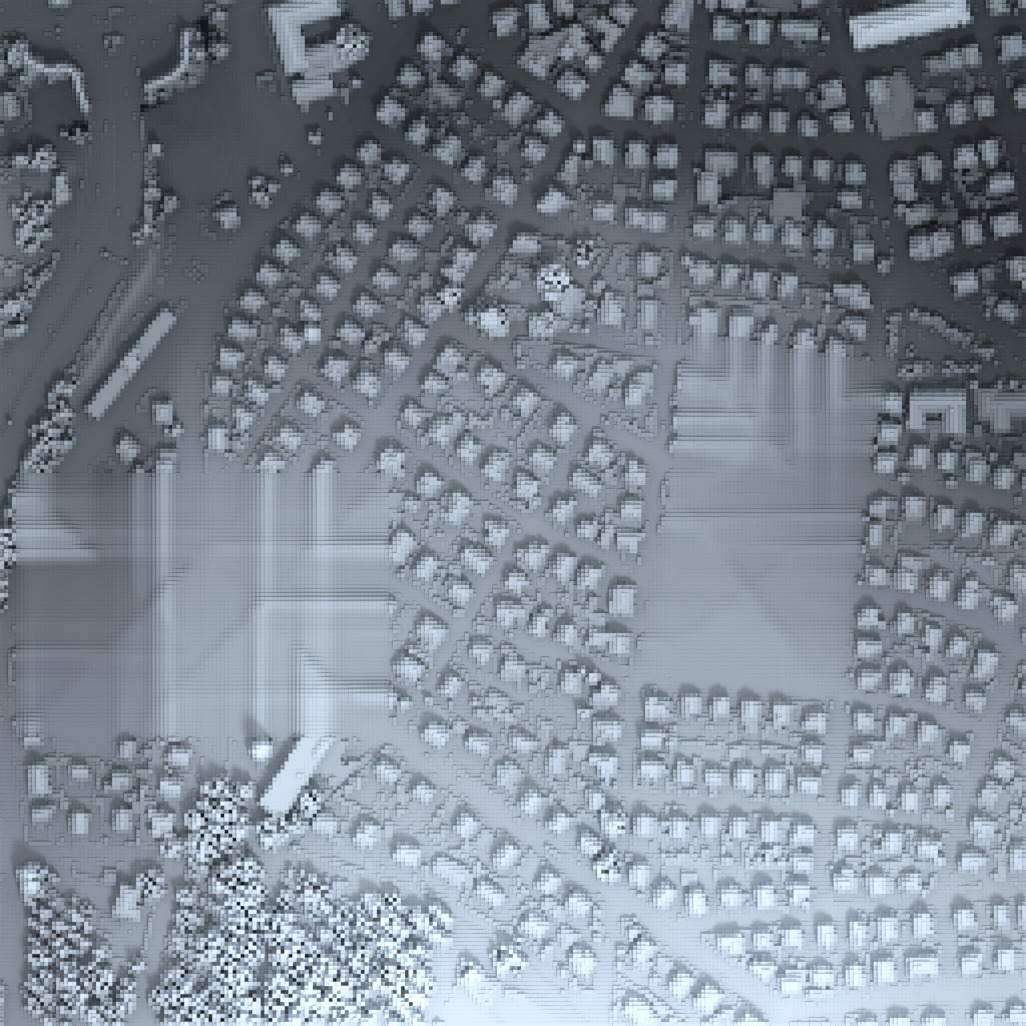} &
\includegraphics[frame, width = 0.18\textwidth]{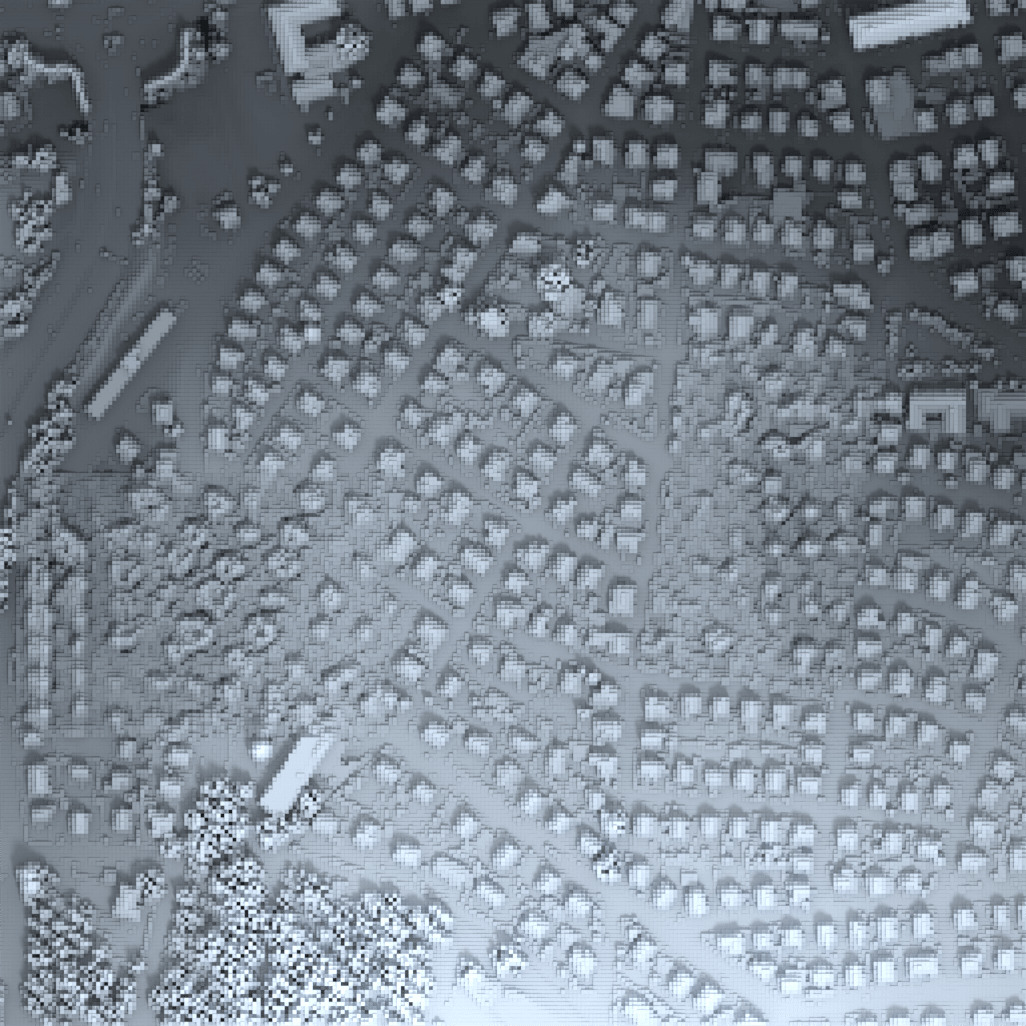}\\

\begin{tikzpicture}
  \node[anchor=south west,inner sep=0] (image) at (0,0) {\includegraphics[frame, width = 0.18\textwidth]{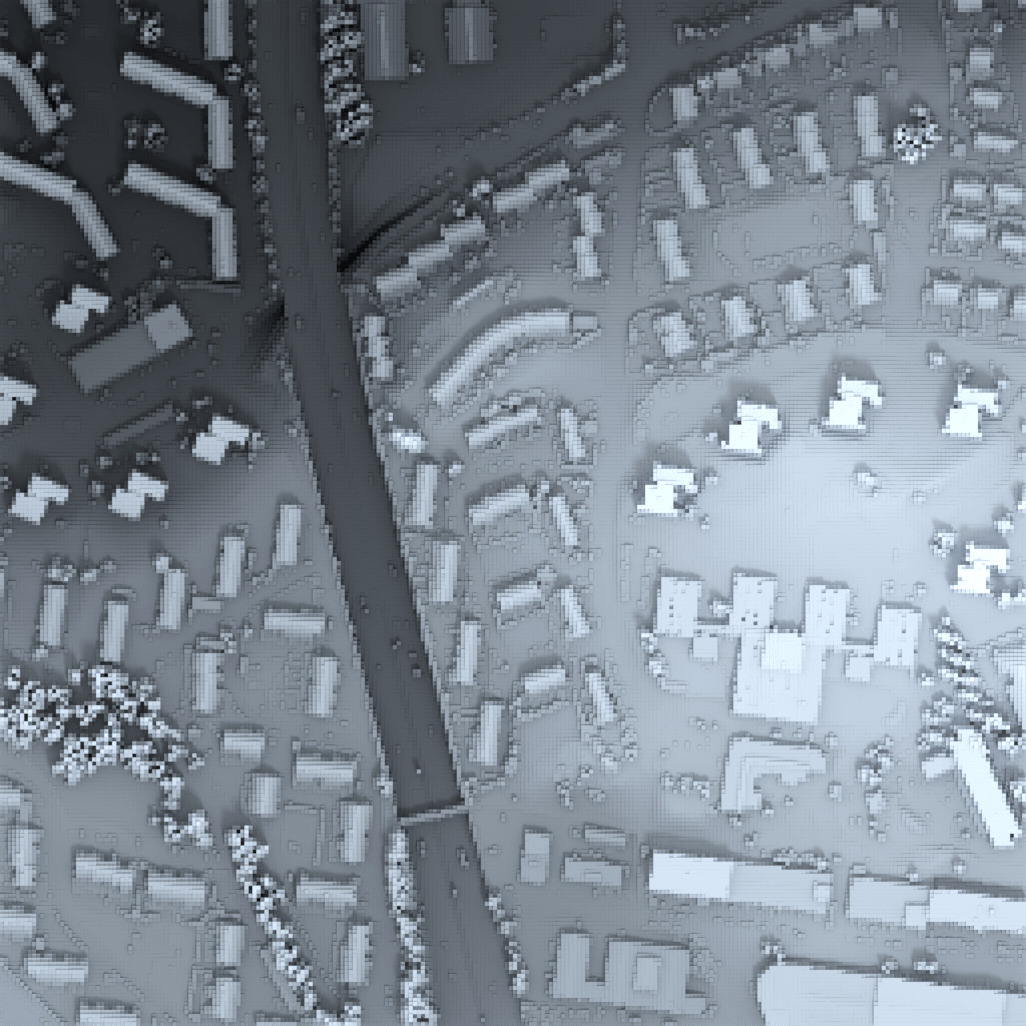}};
  \begin{scope}[x={(image.south east)},y={(image.north west)}]

  \node[above right] at (0.01,0.01) {\footnotesize\textbf{(f)}};

  \end{scope}
\end{tikzpicture} &
\includegraphics[frame, width = 0.18\textwidth]{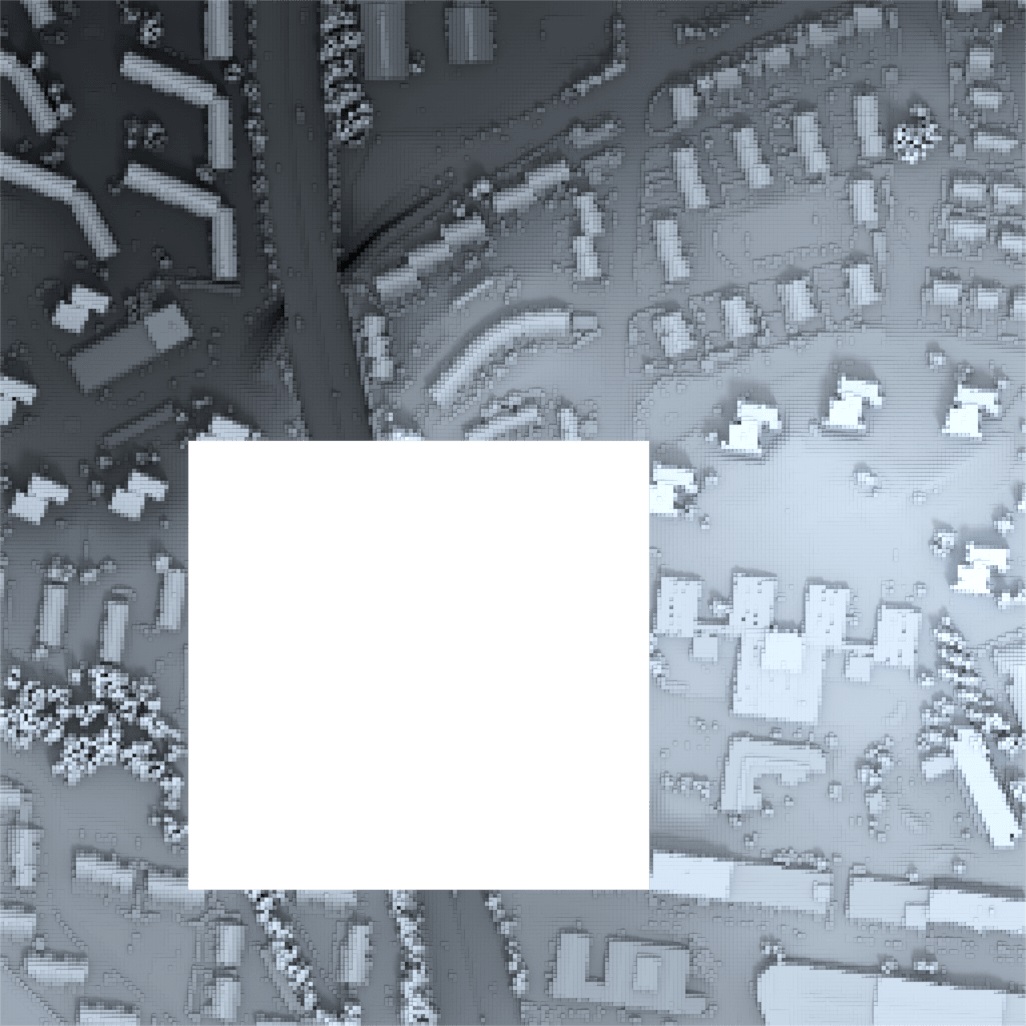} &
\includegraphics[frame, width = 0.18\textwidth]{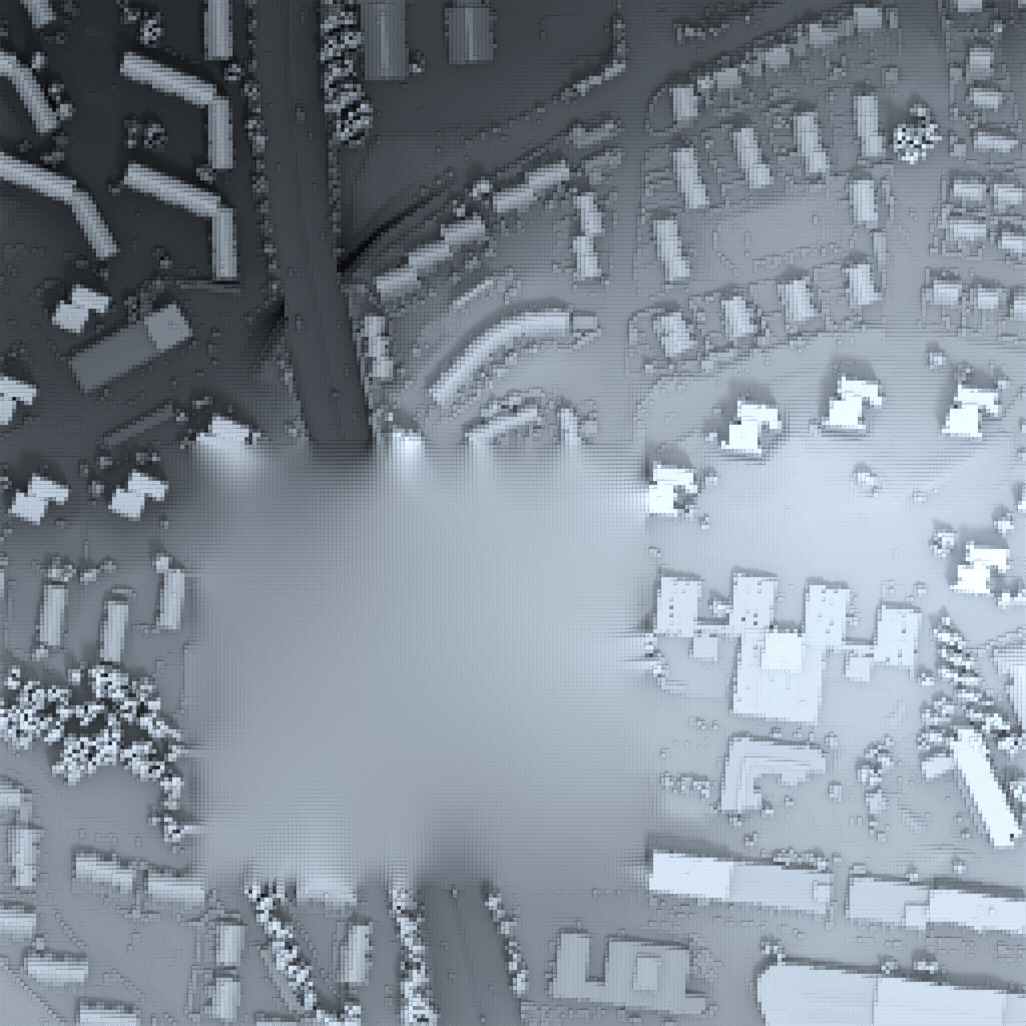} &
\includegraphics[frame, width = 0.18\textwidth]{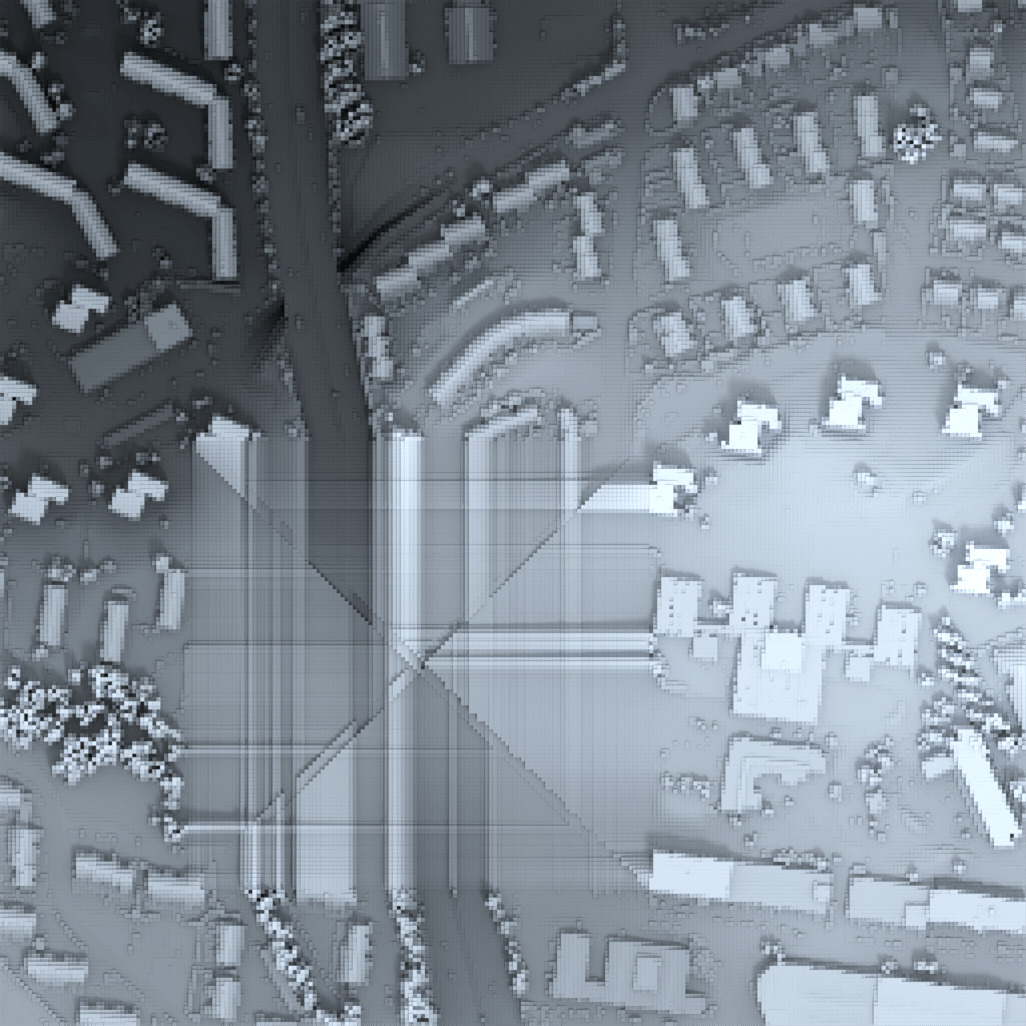} &
\includegraphics[frame, width = 0.18\textwidth]{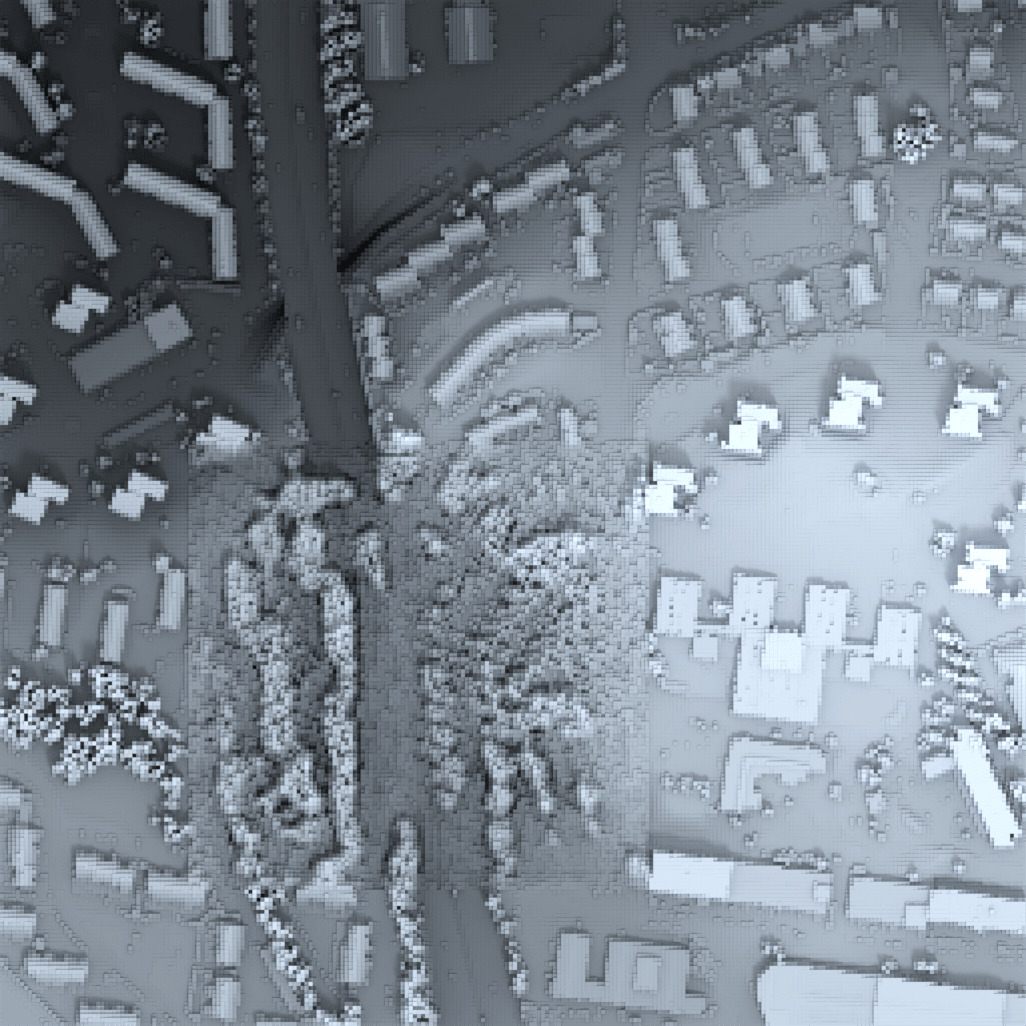}

\end{tabularx}

\vspace{10pt}
\noindent\hrulefill
\caption{A selection of results of our application to (a)--(c) rural and (d)--(f) urban data, rendered such that shadows emphasize any artifacts in the image. From left to right: original DEM, mask, LR B-spline approximation, IDW, our generator $G$. Row (f) shows a failure case, in that it fails to reconstruct the road forming the most prominent feature.
\label{fig: examples}}
\end{figure*}

Figure \ref{fig: examples} contains a carefully selected collection of representative scenarios. These include large missing regions (Figures \ref{fig: examples}\{a,d\}), multiple voids (Figures \ref{fig: examples}\{b,c,e\}), and non-axis-aligned voids (Figure \ref{fig: examples}e). The strength of the spline method is to smoothly interpolate the boundary of the missing regions. The IDW method gives good results on small gaps, but shows axis-aligned and diagonal artifacts on larger grids. Our approach typically achieves the expected geometric continuation and respects surrounding features.


The generator $G$, IDW, and LR B-spline methods were also applied to randomly selected urban and rural DEMs, 50 of each. The results were compared to the ground truth in the EM distance applied to histograms of intensities, and the MSE, as summarized in Table~\ref{tab:QuantitativeComparison}. Complete recovery of the ground truth is an unrealistic goal, so these results provide limited insight. However, quantitative measures, while being less subjective, may not directly correspond to how humans perceive and judge generated images \cite{Borji2018}, which is reflected in the table.

\begin{table}[H]
\caption{Average MSE and EM errors for the various void filling methods, with the generator $G_1$, $G_2$ trained on the urban, rural datasets respectively.}\label{tab:QuantitativeComparison}
\begin{tabular*}{\columnwidth}{@{ }ll@{\extracolsep{\stretch{1}}}*{4}{r}@{ }}
\toprule
  & & $G_1$ & $G_2$ & IDW & LR B-spline\ \\
 \midrule
Urban & MSE & $28.76$ & - & $22.10$ & $\mathbf{17.08}$\ \\
      &  EM & $10.28$ & - & $\mathbf{8.21}$ & $12.99$\ \\ \midrule
Rural & MSE & - & $\mathbf{809.09}$ & $1079.01$ & $868.13$\ \\
      &  EM & - & $\mathbf{8.55}$ & $9.37$ & $11.32$\ \\
\bottomrule
\end{tabular*}
\end{table}


\section{Conclusion}

In this letter, we adapt an existing methodology for image inpainting to filling voids in a DEM. We present results from multiple usage scenarios and showcase the advantages and the drawbacks of our approach.

We consider this work as a generic proof of concept, establishing viability of using deep generative models in the context of DEMs. As such we have limited the scope of the presented methodology to the task of filling missing regions in DEMs. However, we identify the wider applicability of our pre-trained model to other types of remote sensing data (domain adaptation) and related tasks (transfer learning), such as manipulating existing data. By making the model and other resources publicly available \cite{DEM-fill}, we encourage the reader to transfer these results to related applications domains.

In the future we would like to extend this methodology to targeted applications, such as superresolution and generating prescribed structures by replacing input noise vectors by interpretable code vectors. The latter can be achieved by disentangling the individual entries of the code vector by introducing a component that minimizes mutual information \cite{Chen2016InfoGAN}. Another natural next step is multi-view learning for consolidating various data sources, for instance by stacking these views as separate input layers. Multi-task learning holds a high potential for extracting more generic features that are more suitable for transfer learning to specific tasks. Finally we wish to investigate using other evaluation metrics \cite{Borji2018} more suitable to this use case.

\section*{Acknowledgments}

We adapted a publicly available GitHub repository for our experiments \cite{Yu2018}.
The open source C++ library GoTools was also used for generating the LR B-spline data \cite{GoTools2018}.
Data provided courtesy Norwegian Mapping Authorities (www.hoydedata.no), copyright Kartverket (CC BY 4.0).


\ifCLASSOPTIONcaptionsoff
  \newpage
\fi

\bibliographystyle{IEEEtran}
\bibliography{./bibtex/IEEEabrv,./bibtex/bib}

\end{document}